\documentclass[two column]{svjour3}
\usepackage[noadjust]{cite}
\usepackage{amssymb,amsmath}
\usepackage{soul}
\usepackage{graphics}
\usepackage{psfig}
\usepackage{epsfig}
\usepackage{graphicx}
\usepackage{algorithmic}
\usepackage[ruled]{algorithm2e}
\usepackage{morefloats}
\usepackage{color,soul}
\usepackage[most]{tcolorbox}
\usepackage{url}
\usepackage{mathtools, cuted}

\begin{document}

\title{Deep Neural Networks for Automatic Grain-matrix Segmentation in Plane and Cross-polarized Sandstone Photomicrographs}

\author{Rajdeep Das$^{1}$, Ajoy Mondal$^{2}$, Tapan Chakraborty$^{1}$, and Kuntal Ghosh$^{1}$}
\institute{$^{1}$ Indian Statistical Institute, Kolkata, India \\
Email: rajdeep0129@gmail.com, tapan.gsu@gmail.com and kuntal@isical.ac.in \\
$^{2}$ CVIT, International Institute of Information Technology, Hyderabad, India \\ Email: ajoy.mondal83@gmail.com}
\maketitle

\begin{abstract}

Grain segmentation of sandstone that is partitioning the grain from its surrounding matrix/cement in the thin section is the primary step for computer-aided mineral identification and sandstone classification. The microscopic images of sandstone contain many mineral grains and their surrounding matrix/cement. The distinction between adjacent grains and the matrix is often ambiguous, making grain segmentation difficult. Various solutions exist in literature to handle these problems; however, they are not robust against sandstone petrography's varied pattern. In this paper, we formulate grain segmentation as a pixel-wise two-class (i.e., grain and background) semantic segmentation task. We develop a deep learning-based end-to-end trainable framework named Deep Semantic Grain Segmentation network ({\sc dsgsn}), a data-driven method, and provide a generic solution. As per the authors' knowledge, this is the first work where the deep neural network is explored to solve the grain segmentation problem. Extensive experiments on microscopic images highlight that our method obtains better segmentation accuracy than various segmentation architectures with more parameters. 

\end{abstract}

\section{Introduction}

Image segmentation is a basic and crucial step in many image analysis and computer vision tasks. The task of image segmentation sometimes reduces to the form of binary classification, which is often referred to as figure-ground or object-background segmentation. Object-background segmentation has many important applications like document analysis~\cite{vo2018binarization,he2019deepotsu}, visual searching~\cite{wolfe2002segmentation}, saliency detection~\cite{hou2007saliency}, and video tracking~\cite{chan2018segmentation}. In segmentation of the objects from the background, the goal is to divide the image into two homogeneous partitions based on luminance, color, texture, etc. Choosing the appropriate threshold in image histograms is often an effective method for such object-background partitioning. However, this simple binary classification task can turn out to be intriguing, especially in situations involving uncertainties and ambiguities with color, luminance and/or texture information, or when multiple objects are to be detected from their background and such objects have overlapping boundaries. The problem becomes complicated when both these situations mentioned above are present in the image, and the segmentation methods cited above become inadequate.  The sandstone photomicrographs captured under petrographic microscope present one such typical situation where the Geologists still are heavily dependent on tedious manual segmentation for identification of grains from matrix~\cite{dias2020petrographic}, the details of which are explained below.

Sandstones that makeup about one-fourth to one-fifth of all sedimentary rocks are clastic sedimentary rocks composed mainly of sand-sized mineral particles or rock fragments (clasts) and matrix which are finer-grained sedimentary material, such as clay and silt that surround or occur in the interstices of larger grains. Like any other sedimentary rock, these are deposited layer by layer, forming what is usually referred to as bedding. The internal features of the sandstone are indicative of the surface processes and climatic conditions prevailing in the past. Its study involving fossil records also leads to a deeper understanding of the evolution of life on earth. Assessing the relative proportion of the grain and matrix, distribution of the grain size, determining the grain shape and its mineralogy need to be routinely performed in order to interpret the significance of the sandstone. The segmentation of grains in a sandstone thin-section under the plane and cross-polarized light in a microscope is a fundamental step in this study. The grains are the larger fragments to be segregated from the matrix (the finer-grained sedimentary material, such as clay, silt or cementing material that surrounds or occurs in the interstices of larger grains) of the rocks. The geologist usually identifies the grains by studying a thin section observed under a polarizing microscope or a photomicrograph image of the thin section. 
 
A photomicrograph typically contains hundreds of mineral grains in which the manual segmentation is very tedious, time-consuming, and subjective to individual expertise. This is especially so, in view of the overlapping grain boundaries in the two-dimensional images, as well as the large scale variation of color and intensity of the grains/clasts depending upon several factors like the crystal structure of the minerals, orientation of the crystal axes with respect to the plane of microscope stage, thickness of mineral grains and other optical properties of the grain. In plane polarised light in a petrographic microscope (only with the basal polariser used) one can see certain properties of the grains like refractive index, body colour etc., whereas in cross-polar view (with both the polariser and the analyser inserted), one can observe properties like birefringence, extinction position etc. In case of isotropic materials (like glass, water, mineral garnet), under cross polar view, all the light transmitted through the mineral are blocked by the analyser and the grain appears dark in all position of the rotating microscope stage. However, in case of anisotropic minerals, in cross-polar view one can observe light of different brightness and colour in different grains of the same mineral (due to variation in the orientation of the crystals in different grains). As the microscope stage is rotated, the brightness changes continuously. Thus an anisotropic mineral grain (quartz, feldspar etc) in one position may show bright grey colour while in the other position it may be completely dark.  Further, due to various other processes operating during the transformation of loose sand to sandstone, the boundaries of grains can be modified to be extremely complex and often fuzzy, hindering easy delineation of the grain boundary from surrounding matrix. This makes the task of automated grain segmentation even more challenging. The photomicrographs meant for such automated segmentation, depending upon the orientation and type of the various crystals present in the slides,  display wide variance in grain luminance from almost invisibly dark to very bright in appearances. Yet, such apparent variation may in fact often be misleading, a fact which makes automated grain segmentation in sandstone photomicrographs an all the more difficult task. This is why, in absence of reliable computer vision algorithms for grain segmentation (despite several important attempts mentioned in the following section), the geologists have to combine all the features observable in plane and cross-polarised view and in different positions of the rotating microscope stage, and have to perform the segmentation task manually and tediously.  

This paper poses automatic grain-matrix segmentation of plane and cross-polarized microscopic sandstone images as pixel-wise two-class semantic segmentation tasks to deal with this challenging problem. For this purpose, we develop a deep learning-based end-to-end trainable framework named Deep Semantic Grain Segmentation Network ({\sc dsgsn}), inspired by the LinkNet architecture~\cite{linknet}. The \textsc{dsgsn} consists of an encoder network and a corresponding decoder network followed by a pixel-wise classification layer. The developed network takes sandstone images as input and produces segmented images as output. The proposed network is data-driven and learns features from the training set during the training period. The trained model has a generic solution during inference time. Usually, deep learning-based approaches require a large amount of labeled training data which is not available for this particular problem. We apply various data augmentation schemes to solve the scarcity of labeled training data for this current work. Due to the insufficient availability of such images with ground truth annotations, we create a dataset consisting of seven pairs of images with pixel-level ground truth annotations. Experiments on the newly generated dataset conclude that the proposed approach is superior to the various techniques for segmenting sandstone images.    
We summarize the contribution of this work as follows.

\begin{itemize}
\item We formulate Grain-matrix segmentation as a pixel-wise two-class semantic segmentation task and propose a deep learning-based end-to-end trainable framework named Deep Semantic Grain Segmentation Network ({\sc dsgsn}) to solve this task.
\item We create a dataset, named as {\sc isi}-microscopic sandstone image, which consists of $7$ pair of images with manually annotated ground truths.
\item Extensive experiments conclude that the proposed method obtains better results. 
\end{itemize}

The rest of the paper is organized as follows. 
Section~\ref{related_work} the related works in the literature corresponding to grain segmentation or grain boundary detection in rock thin section images.
Section~\ref{proposed_method} gives a brief description of deep semantic grain segmentation approach. We analyze the obtained results in Section~\ref{experiments}. Finally, Section~\ref{conclusion} concludes the present work.

\begin{figure*}[ht!]
\centerline{
\tcbox[sharp corners, size = tight, boxrule=0.2mm, colframe=black, colback=white]{\psfig{figure=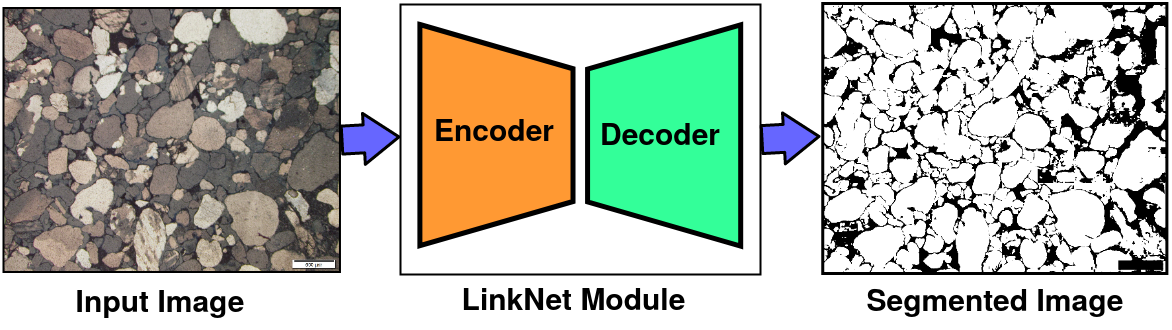, width=1\textwidth,height=0.25\textwidth}}}
\caption{Visually illustrates an overview of the proposed {\sc dsgsn}. It takes an input image and produces a segmented image as output. The network's encoder learns the features while the decoder reconstructs the foreground image by up-sampling the learned features for classification.\label{fig:block_diagram}}
\end{figure*}

\section{Related Work} \label{related_work}

\subsection{Grain Segmentation}

Significant advances have been made in the domain of computational technologies on one hand and image processing and analysis techniques, on the other, especially in the last three decades. Several efforts have been made to deal with the problem of automatic grain segmentation of rock thin section images. The task of grain-matrix segmentation has been approached effectively through two alternative paths: grain boundary detection and direct segmentation of the grains from their background, i.e., the less-significant matrix zone in the rock thin section. However, the main objective for the approaches is the same, viz. the identification of grains. As distinct from the early attempts to detect grain boundary in igneous and metamorphic rocks~\cite{lumbreras1996segmentation}, in which the matrix portion is almost negligible, the task in sedimentary rocks is much more critical because of the heterogeneous form of organization in grain and matrix structure and uncertain boundaries between them at places. In~\cite{lumbreras1996segmentation}, the authors introduce an unsupervised segmentation procedure for petrographic images of marbles by first applying watershed transform~\cite{vincent1991watersheds} on the images and then by merging the over-segmented regions using their pixel-wise module and phase parameter values which is a characteristic of each grain. In~\cite{goodchild1998edge}, the authors propose a simple gradient-based edge detection technique, much like Canny's edge detection paradigm~\cite{canny1986computational}. In 2002, Berg {\em et al.} devised a technique~\cite{van2002automated} for separating touching grains in rock thin section images and compared their result with that produced with simple watershed transform~\cite{vincent1991watersheds}. Zhou {\em et al.} in~\cite{zhou2004segmentation} segmented petrographic images by integrating edge detection and region growing. Unlike in most of the previous papers and the approach adopted in our present work, segmentation has been performed based on a single view (image captured in cross-polar mode only at a fixed orientation of the stage of the petrographic microscope) rock thin section image. The authors have first mathematically formulated a gradient-based variable size filter for probable grain boundary edge (including inner grain texture edge) detection and then used seeded region growing procedures generating seeds at categorized regions and finally merged judiciously the candidate regions to arrive at the final segmentation result. In a similar work, Smith and Beermann in~\cite{smith2007image} detected plagioclase crystal boundaries in thin-section images of rocks by using grey level homogeneity recognition of discrete areas. In~\cite{fueten2007artificial} Fueten and Mason have used the same edge detection technique as used in~\cite{goodchild1998edge} previously and edited edges with the assistance of a three-layered feed-forward neural network using feature values on pixel intensities and texture of grain portion on both sides of earlier detected edges to eliminate false edges in a supervised mode. The authors of the paper~\cite{lu2009automated} applied partial differential equation-based segmentation by level set formulation as in~\cite{chan2001active} to segment nearly equal intensity grains from background matrix in the color micrograph of thin sections. A cellular automata~\cite{chang2004cellular} based approach was also investigated~\cite{gorsevski2012detecting} for automatic detection of grain boundaries in deformed rocks. Here the authors used a set of images of thin sections of sandstone taken at multiple angles of the petrographic microscope for better clarification of the problem. In~\cite{yesiloglu2012computer} the authors applied a method called TSecSoft. They at first used {\sc jseg} image segmentation algorithm~\cite{dengy2001unsupervisedseg} for segmenting sample rock thin section images on an approximation basis and then interactively edited the segmentation results to determine overall mineral percentage in the sample. A graph-based clustering algorithm was applied in~\cite{mingireanov2013segmentation}. Here firstly, the touching grains in rock thin section images were separated with the help of optimum path forest method~\cite{falcao2004image}. Secondly, the result's editing was effected using live marker method~\cite{spina2011user} with minor changes shown to have occurred. A region competition and edge-weighted region merging based method was used in~\cite{jungmann2014segmentation} utilizing the popular region competition method of~\cite{zhu1996region} and~\cite{mansouri2006multiregion}. The grain segmentation problem is an unknown class clustering or segmentation problem. So, normal K-means clustering with color intensities or texture feature values may not be applied effectively as it necessitates the number of clusters as a parameter. Judiciously, to avoid this problem, an incremental color intensity-based clustering method was employed in~\cite{izadi2015new} for segmenting igneous rock thin section images containing altered minerals as well as those not containing altered minerals. In very recent work, in~\cite{izadi2020altered}, the authors have provided an improved, and better solution than methods used in~\cite{izadi2015new} and~\cite{izadi2017intelligent} producing segmentation results for thin section images with non-altered and specifically altered minerals with higher accuracy by applying both incremental and dynamic clustering approach. In another recent work, Maitree {\em et al.}~\cite{maitre2019mineral} have effectively used the Super-pixel based method of Simple Linear Iterative Clustering ({\sc slic}) to achieve encouraging results. The Super-pixel method has also been used by Jiang {\em et al.}~\cite{jiang2017grain,jiang2018method} among other recent important works.

\subsection{Image Semantic Segmentation using Deep Learning} \label{image_semantic_segmentation}

Pixel-wise classification task like image semantic segmentation achieves significant improvement inspired by replacing the fully connected layer in the image classification network with the convolution layer, referred to as Fully Convolution Network ({\sc fcn})~\cite{long2015fully}. Limitations of the {\sc fcn} model~\cite{long2015fully} include ignoring small objects and mislabeling the large objects due to fixed receptive field size. To address this issue, Noh {\em et al.}~\cite{noh2015learning} proposed a semantic segmentation algorithm by learning a deconvolution network composed of both deconvolution and unpooling layers. Several methods~\cite{chen2018deeplab,zhao2017pyramid} used dilated convolution to enlarge the receptive field of the neural network. Since in deep networks, higher-layer feature contains more semantic meaning and less location information, various methods~\cite{zhao2017pyramid,chen2016attention} have been developed by combining multi-scale features to improve the segmentation performance. Another direction in semantic segmentation is based on structure prediction. The pioneer work~\cite{chen2018deeplab} considers the conditional random field ({\sc crf})~\cite{crf} for post-processing to refine the segmentation result. Various methods~\cite{arnab2016higher,zheng2015conditional} included {\sc crf} to refine the networks' end-to-end modeling. The inclusion of {\sc crf} improves the segmentation boundaries.

All these above discussed networks consist of encoders and their corresponding decoders. Some spatial information is lost in decoders because of performing multiple down-sampling operations (max-pooling) in the encoders. It is very difficult to recover this lost information in decoders using only the down-sampled outputs of encoders. Badrinarayanan {\em et al.}~\cite{badrinarayanan2017segnet} solved this problem by linking encoder with decoder through pooling indices which are not trainable parameters. In this direction, various architectures such as {\sc u-n}et~\cite{ronneberger2015u} and {\sc u-n}et++~\cite{zhou2018unet++} with trainable parameters have been proposed to solve this problem. However, all these discussed networks use {\sc vgg}$16$ ($138$ million parameters) or {\sc r}es{\sc n}et$101$ ($45$ million parameters) as their encoder which are huge in terms of parameters and {\sc gflop}s, excepting {\sc u-n}et++. These architectures can not be used for real-time applications. Various networks such as {\sc erf}net~\cite{romera2017erfnet} and {\sc l}ink{\sc n}et~\cite{linknet} are developed for real-time applications.

\section{Deep Semantic Grain Segmentation}\label{proposed_method}

While visual recognition has been an active research topic for the last few decades, there has been a recent trend shift in its research from traditional hand-crafted feature design, as discussed in the previous section, to feature extraction through deep networks. Through several instances in computer vision literature, it has recently been proven that the latter data-driven approach often
far outperforms the hand-crafted feature-based earlier approaches. We pose grain segmentation as a pixel-wise two-class (i.e., grain and background) semantic segmentation task in this work. For this purpose, we develop a deep learning-based end-to-end trainable framework named Deep Semantic Grain Segmentation Network ({\sc dsgsn}), inspired by the LinkNet architecture~\cite{linknet}. The \textsc{dsgsn} consists of an encoder network and a corresponding decoder network followed by a pixel-wise classification layer. It takes an original image, $x$ as input, and produces a segmented image $y$ as output. Figure~\ref{fig:block_diagram} illustrates the overview of the \textsc{dsgsn}. Let, $x_{ij}$ represent the information of the input image $x$ at pixel location $(i,\ j)$, while $y_{ij}$ represents the class label (i.e. $1$ for grain and $0$ for background) at pixel position $(i,\ j)$ of the output segmented image $y$.

Let the ground truth segmented image be $y^{gt}$ corresponding to the input image $x$. Therefore, $y_{ij}^{gt}$ denotes the class label (i.e., $1$ for grain and $0$ for background) at pixel location $(i,\ j)$ of the ground truth segmented image $y^{gt}$. We formulate the semantic segmentation task as $y = f(x)$, where $x$ is the input image, $y$ is the segmented image and $f$ is the pixel-wise non-linear mapping function. During training, given a pair of images $(x,\ y^{gt})$, the network learns this mapping function to the network's weight parameters. The mapping function $f$ is represented with the network's weight parameters learned by minimizing pixel-wise weighted two-class cross-entropy loss between the predicted segmented image $y$ and ground truth segmented image $y^{gt}$.  We define the loss function as

\begin{equation}
 E = \sum_{i}\sum_{j} -w_{1}  y_{ij} log(y_{ij}^{gt}) - w_{0} (1-y_{ij})\ log(1-y_{ij}^{gt}), \label{loss_function}
\end{equation}
where $w_{1} = \frac{N_{obj}}{N}$ and $w_{0}=\frac{N_{bg}}{N}$ with $N_{obj}$, $N_{bg}$ and $N$ are the number of pixels within the grain region, number of pixels within the background region and the total number of pixels in an image of the training set, respectively.

\begin{figure}[ht!]
\centerline{
\tcbox[sharp corners, size = tight, boxrule=0.2mm, colframe=black, colback=white]{
\psfig{figure=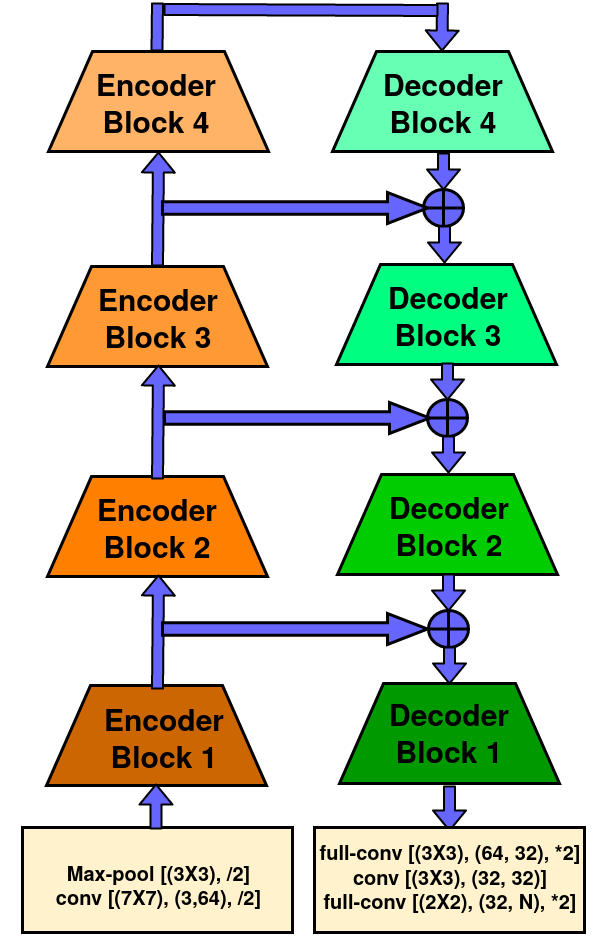, width=0.4\textwidth,height=0.5\textwidth}}}
\caption{Shows the encoder-decoder architecture in the LinkNet module. \textbf{Left:} shows the convolution modules in the encoder network, and \textbf{Right:} shows the convolution modules in the decoder network.\label{fig:architecture}}
\end{figure}

Recently, deep semantic architectures~\cite{long2015fully,ronneberger2015u,chen2017deeplab,badrinarayanan2017segnet,zhao2017pyramid} have been developed in the literature to segment images of natural scenes. All these methods used \textsc{vgg}-$16$ ($138$ million parameters)~\cite{simonyan2014very} and ResNet$101$ ($45$ million parameters)~\cite{he2016deep} as encoder with huge number of parameters. Due to such a large number of parameters, the amount of training images required to fine-tune the network for doing specific tasks is large. In the case of grain segmentation, it is challenging and hardly cost-effective to build such a large, manually annotated dataset. Due to the multiple down-sampling operations in the encoder, spatial information is also lost. It is challenging to recover the spatial information lost in the encoder, using multiple up-sampling operations in the decoder. Hence, there is a need to preserve the input image's semantic structure and pass it on to the up-sampling layers without increasing any computational cost, in order to overcome the problem mentioned above.

Therefore, our objective is to use a network that tries to solve this problem by keeping the network size small while producing better results with a limited number of training samples. To this end, our designed \textsc{dsgsn} inspired by the LinkNet architecture~\cite{linknet} attains the defined objective.   

\subsection{Network Architecture}

\begin{figure}[ht!]
\centerline{
\tcbox[sharp corners, size = tight, boxrule=0.2mm, colframe=black, colback=white]{
\psfig{figure=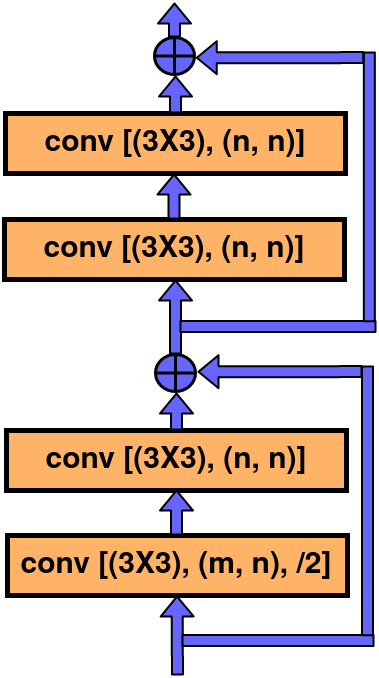, width=0.4\textwidth,height=0.5\textwidth}}}
\caption{Shows the details of convolution modules in the encoder-block(i). \label{fig:residual}}
\end{figure}

\begin{figure}[ht!]
\centerline{
\tcbox[sharp corners, size = tight, boxrule=0.2mm, colframe=black, colback=white]{
\psfig{figure=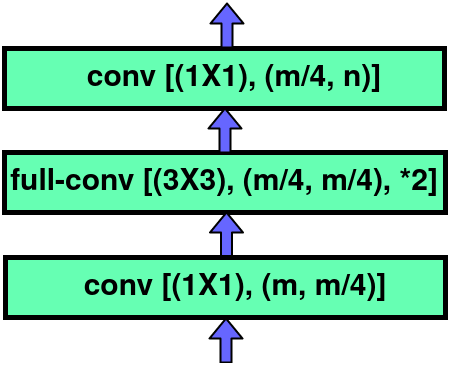, width=0.3\textwidth,height=0.35\textwidth}}}
\caption{Shows the details of convolution modules in the decoder-block(i). \label{fig:decoder}}
\end{figure}

Figure~\ref{fig:architecture} displays our encoder-decoder architecture. In this figure, `Conv' means convolution, `Full-conv' means full convolution, and `/$2$' indicates down-sampling by a factor $2$, which is performed by stride convolution. Similarly, `*$2$' represents up-sampling by a factor of $2$. We consider batch normalization between the convolutional layer and ReLU. The encoder-decoder network in Figure~\ref{fig:architecture} indicates the left part as an encoder and the right part as a decoder. 
The encoder and the decoder, in turn, consist of several encoder and decoder blocks, respectively, as shown in Figure~\ref{fig:architecture}. The initial encoder block performs convolution on the input image with $7 \times 7$ kernel and a stride $2$. After that, max-pooling is performed on $3 \times 3$ area with stride $2$. All other encoder blocks consist of a residual block~\cite{he2016deep}, which is represented as encoder-block(i). Individual layers within these blocks are shown in Figure~\ref{fig:residual}. Similarly, the detailed layers in the decoder-blocks are provided in Figure~\ref{fig:decoder}. We consider ResNet$18$ ($11.5$ million parameters)~\cite{he2016deep}, a lighter network as compared to \textsc{vgg}-$16$ and ResNet$101$, as an encoder. Fine-tuning this network for a specific task with the limited training samples is not a problem. On the other hand, the information lost in the encoder is retained by bypassing the output of each encoder to its corresponding decoder, which is shown in Figure~\ref{fig:architecture}. In this architecture, the decoder is sharing the knowledge learned by the encoder at every layer. The decoder uses fewer parameters resulting in an efficient network.

\begin{figure}[ht!]
\centerline{
\tcbox[sharp corners, size = tight, boxrule=0.2mm, colframe=black, colback=white]{
\psfig{figure=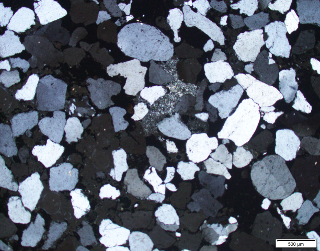, width=0.15\textwidth,height=0.12\textwidth}}
\hspace{-0.005\textwidth}
\tcbox[sharp corners, size = tight, boxrule=0.2mm, colframe=black, colback=white]{
\psfig{figure=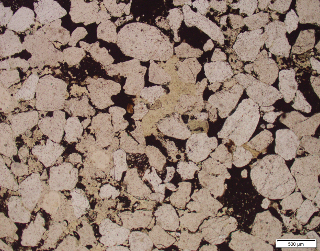, width=0.15\textwidth,height=0.12\textwidth}}
\hspace{-0.005\textwidth}
\tcbox[sharp corners, size = tight, boxrule=0.2mm, colframe=black, colback=white]{
\psfig{figure=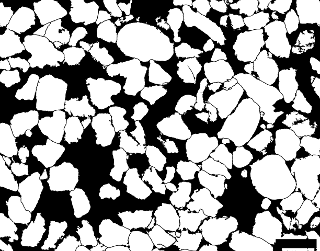, width=0.15\textwidth,height=0.12\textwidth}}}
\vspace{-0.01\textwidth}
\centerline{
\tcbox[sharp corners, size = tight, boxrule=0.2mm, colframe=black, colback=white]{
\psfig{figure=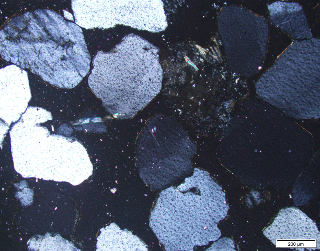, width=0.15\textwidth,height=0.12\textwidth}}
\hspace{-0.005\textwidth}
\tcbox[sharp corners, size = tight, boxrule=0.2mm, colframe=black, colback=white]{
\psfig{figure=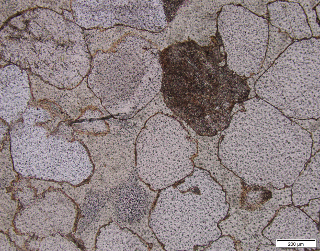, width=0.15\textwidth,height=0.12\textwidth}}
\hspace{-0.005\textwidth}
\tcbox[sharp corners, size = tight, boxrule=0.2mm, colframe=black, colback=white]{
\psfig{figure=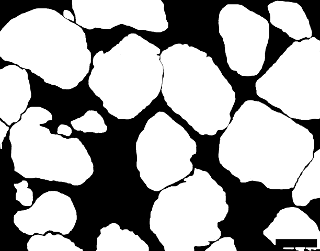, width=0.15\textwidth,height=0.12\textwidth}}}
\vspace{-0.01\textwidth}
\centerline{
\tcbox[sharp corners, size = tight, boxrule=0.2mm, colframe=black, colback=white]{
\psfig{figure=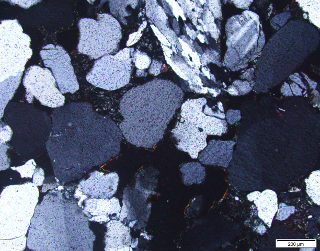, width=0.15\textwidth,height=0.12\textwidth}}
\hspace{-0.005\textwidth}
\tcbox[sharp corners, size = tight, boxrule=0.2mm, colframe=black, colback=white]{
\psfig{figure=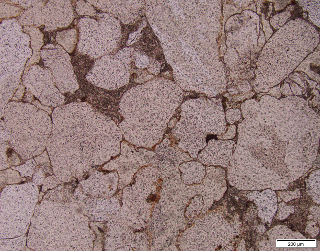, width=0.15\textwidth,height=0.12\textwidth}}
\hspace{-0.005\textwidth}
\tcbox[sharp corners, size = tight, boxrule=0.2mm, colframe=black, colback=white]{
\psfig{figure=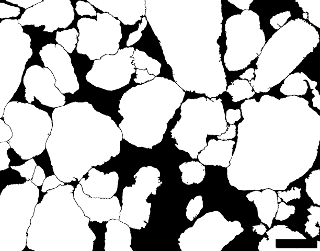, width=0.15\textwidth,height=0.12\textwidth}}}
\vspace{-0.01\textwidth}
\centerline{
\tcbox[sharp corners, size = tight, boxrule=0.2mm, colframe=black, colback=white]{
\psfig{figure=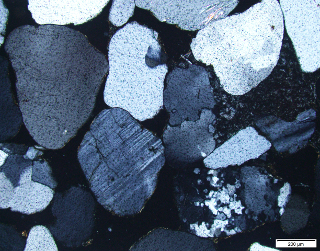, width=0.15\textwidth,height=0.12\textwidth}}
\hspace{-0.005\textwidth}
\tcbox[sharp corners, size = tight, boxrule=0.2mm, colframe=black, colback=white]{
\psfig{figure=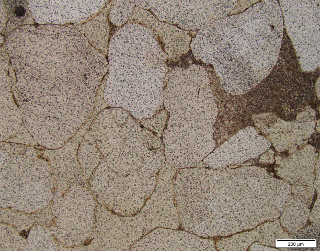, width=0.15\textwidth,height=0.12\textwidth}}
\hspace{-0.005\textwidth}
\tcbox[sharp corners, size = tight, boxrule=0.2mm, colframe=black, colback=white]{
\psfig{figure=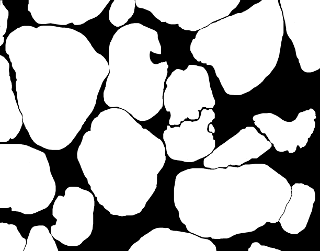, width=0.15\textwidth,height=0.12\textwidth}}}
\vspace{-0.01\textwidth}
\centerline{
\tcbox[sharp corners, size = tight, boxrule=0.2mm, colframe=black, colback=white]{
\psfig{figure=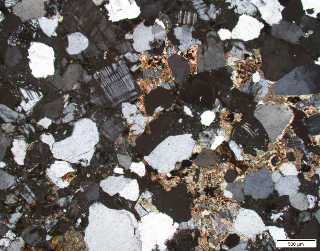, width=0.15\textwidth,height=0.12\textwidth}}
\hspace{-0.005\textwidth}
\tcbox[sharp corners, size = tight, boxrule=0.2mm, colframe=black, colback=white]{
\psfig{figure=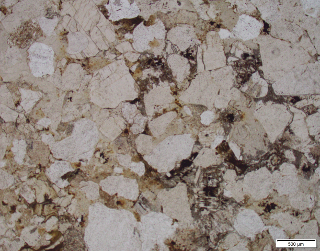, width=0.15\textwidth,height=0.12\textwidth}}
\hspace{-0.005\textwidth}
\tcbox[sharp corners, size = tight, boxrule=0.2mm, colframe=black, colback=white]{
\psfig{figure=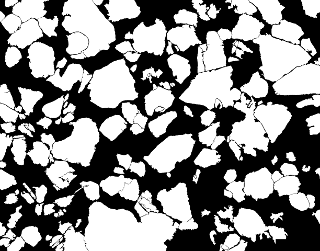, width=0.15\textwidth,height=0.12\textwidth}}}
\vspace{-0.01\textwidth}
\centerline{
\tcbox[sharp corners, size = tight, boxrule=0.2mm, colframe=black, colback=white]{
\psfig{figure=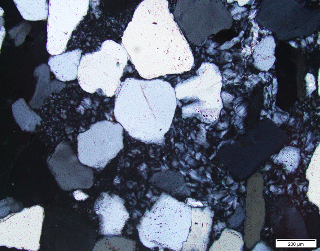, width=0.15\textwidth,height=0.12\textwidth}}
\hspace{-0.005\textwidth}
\tcbox[sharp corners, size = tight, boxrule=0.2mm, colframe=black, colback=white]{
\psfig{figure=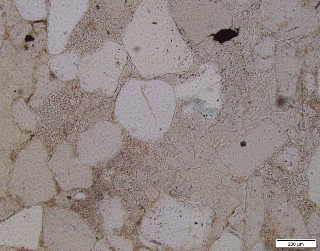, width=0.15\textwidth,height=0.12\textwidth}}
\hspace{-0.005\textwidth}
\tcbox[sharp corners, size = tight, boxrule=0.2mm, colframe=black, colback=white]{
\psfig{figure=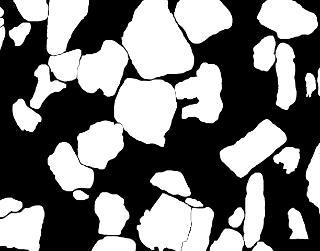, width=0.15\textwidth,height=0.12\textwidth}}}
\vspace{-0.01\textwidth}
\centerline{
\tcbox[sharp corners, size = tight, boxrule=0.2mm, colframe=black, colback=white]{
\psfig{figure=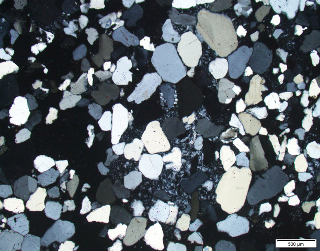, width=0.15\textwidth,height=0.12\textwidth}}
\hspace{-0.005\textwidth}
\tcbox[sharp corners, size = tight, boxrule=0.2mm, colframe=black, colback=white]{
\psfig{figure=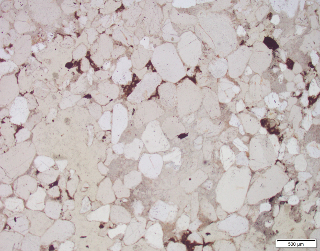, width=0.15\textwidth,height=0.12\textwidth}}
\hspace{-0.005\textwidth}
\tcbox[sharp corners, size = tight, boxrule=0.2mm, colframe=black, colback=white]{
\psfig{figure=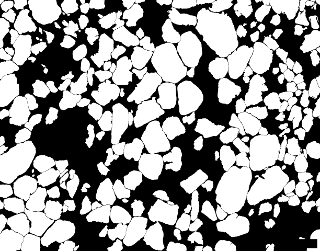, width=0.15\textwidth,height=0.12\textwidth}}}
\caption{Illustrates seven pairs of images with ground truth segmentation in the training set. \textbf{First Column:} shows images in {\sc xpl}. \textbf{Second Column:} shows same images in {\sc ppl} and \textbf{Third Column:} shows corresponding ground truth segmentation. \label{fig:training}}
\end{figure}

\begin{figure}[ht!]
\centerline{
\tcbox[sharp corners, size = tight, boxrule=0.2mm, colframe=black, colback=white]{
\psfig{figure=images/2_ol.png, width=0.15\textwidth,height=0.12\textwidth}}
\hspace{-0.005\textwidth}
\tcbox[sharp corners, size = tight, boxrule=0.2mm, colframe=black, colback=white]{
\psfig{figure=images/3_ol.png, width=0.15\textwidth,height=0.12\textwidth}}
\hspace{-0.005\textwidth}
\tcbox[sharp corners, size = tight, boxrule=0.2mm, colframe=black, colback=white]{
\psfig{figure=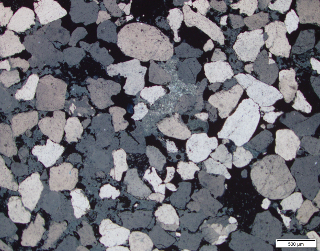, width=0.15\textwidth,height=0.12\textwidth}}}
\vspace{-0.01\textwidth}
\centerline{
\tcbox[sharp corners, size = tight, boxrule=0.2mm, colframe=black, colback=white]{
\psfig{figure=images/6_ol.png, width=0.15\textwidth,height=0.12\textwidth}}
\hspace{-0.005\textwidth}
\tcbox[sharp corners, size = tight, boxrule=0.2mm, colframe=black, colback=white]{
\psfig{figure=images/7_ol.png, width=0.15\textwidth,height=0.12\textwidth}}
\hspace{-0.005\textwidth}
\tcbox[sharp corners, size = tight, boxrule=0.2mm, colframe=black, colback=white]{
\psfig{figure=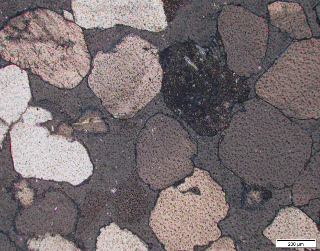, width=0.15\textwidth,height=0.12\textwidth}}}
\vspace{-0.01\textwidth}
\centerline{
\tcbox[sharp corners, size = tight, boxrule=0.2mm, colframe=black, colback=white]{
\psfig{figure=images/8_ol.png, width=0.15\textwidth,height=0.12\textwidth}}
\hspace{-0.005\textwidth}
\tcbox[sharp corners, size = tight, boxrule=0.2mm, colframe=black, colback=white]{
\psfig{figure=images/9_ol.png, width=0.15\textwidth,height=0.12\textwidth}}
\hspace{-0.005\textwidth}
\tcbox[sharp corners, size = tight, boxrule=0.2mm, colframe=black, colback=white]{
\psfig{figure=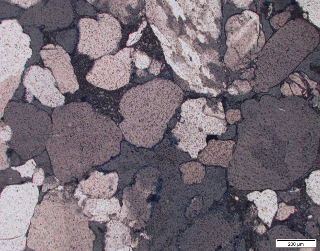, width=0.15\textwidth,height=0.12\textwidth}}}
\vspace{-0.01\textwidth}
\centerline{
\tcbox[sharp corners, size = tight, boxrule=0.2mm, colframe=black, colback=white]{
\psfig{figure=images/11_ol.png, width=0.15\textwidth,height=0.12\textwidth}}
\hspace{-0.005\textwidth}
\tcbox[sharp corners, size = tight, boxrule=0.2mm, colframe=black, colback=white]{
\psfig{figure=images/10_ol.png, width=0.15\textwidth,height=0.12\textwidth}}
\hspace{-0.005\textwidth}
\tcbox[sharp corners, size = tight, boxrule=0.2mm, colframe=black, colback=white]{
\psfig{figure=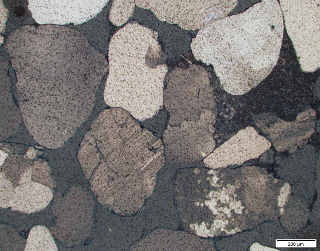, width=0.15\textwidth,height=0.12\textwidth}}}
\vspace{-0.01\textwidth}
\centerline{
\tcbox[sharp corners, size = tight, boxrule=0.2mm, colframe=black, colback=white]{
\psfig{figure=images/14_ol.png, width=0.15\textwidth,height=0.12\textwidth}}
\hspace{-0.005\textwidth}
\tcbox[sharp corners, size = tight, boxrule=0.2mm, colframe=black, colback=white]{
\psfig{figure=images/15_ol.png, width=0.15\textwidth,height=0.12\textwidth}}
\hspace{-0.005\textwidth}
\tcbox[sharp corners, size = tight, boxrule=0.2mm, colframe=black, colback=white]{
\psfig{figure=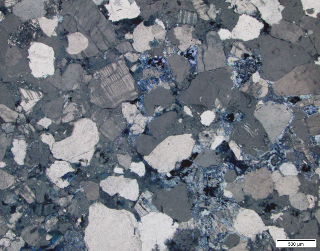, width=0.15\textwidth,height=0.12\textwidth}}}
\vspace{-0.01\textwidth}
\centerline{
\tcbox[sharp corners, size = tight, boxrule=0.2mm, colframe=black, colback=white]{
\psfig{figure=images/16_ol.png, width=0.15\textwidth,height=0.12\textwidth}}
\hspace{-0.005\textwidth}
\tcbox[sharp corners, size = tight, boxrule=0.2mm, colframe=black, colback=white]{
\psfig{figure=images/17_ol.png, width=0.15\textwidth,height=0.12\textwidth}}
\hspace{-0.005\textwidth}
\tcbox[sharp corners, size = tight, boxrule=0.2mm, colframe=black, colback=white]{
\psfig{figure=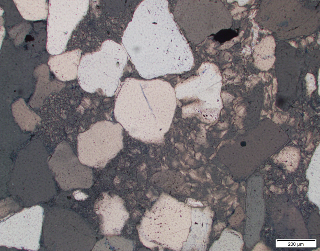, width=0.15\textwidth,height=0.12\textwidth}}}
\vspace{-0.01\textwidth}
\centerline{
\tcbox[sharp corners, size = tight, boxrule=0.2mm, colframe=black, colback=white]{
\psfig{figure=images/20_ol.png, width=0.15\textwidth,height=0.12\textwidth}}
\hspace{-0.005\textwidth}
\tcbox[sharp corners, size = tight, boxrule=0.2mm, colframe=black, colback=white]{
\psfig{figure=images/21_ol.png, width=0.15\textwidth,height=0.12\textwidth}}
\hspace{-0.005\textwidth}
\tcbox[sharp corners, size = tight, boxrule=0.2mm, colframe=black, colback=white]{
\psfig{figure=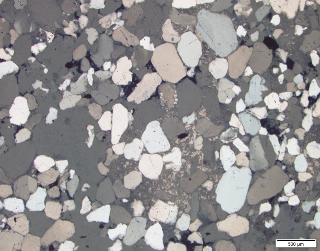, width=0.15\textwidth,height=0.12\textwidth}}}
\caption{Shows the generated average image from a pair of images. \textbf{First Column:} shows images in {\sc xpl}, \textbf{Second Column:} shows images in {\sc ppl} and \textbf{Third Column:} shows the generated average image of \textbf{First Column} and \textbf{Second Column}. \label{fig:average}}
\end{figure}

\section{Experiments \label{experiments}}

\subsection{Dataset}

The sandstone samples used in the current study have been collected by the geologists of the Indian Statistical Institute. Each of the digital images in our dataset represents photographs taken during the sample's thin section microscopic study. The thin sections are slices of about 0.03 mm thickness, mounted on glass slides. A chip of rock is, at first, cut into a small piece. One side is then well-polished and carefully glued to a clean glass slide using a transparent adhesive. The rock chip is finally ground down to the desired final thickness of about 0.03 mm. The images are captured after mounting these slides under the petrographic microscope and then observed in both plane-polar and cross-polar mode. Ten colour photomicrographs taken under both plane (referred in the figures as {\sc ppl} images) and cross-polar (referred in the figures as {\sc xpl} images) light are each of size $1920 \times 2448$ pixels while one sample photomicrograph image set is of size $1536 \times 2048$ pixels. The first-mentioned ten images are captured in `.jpg' format, while the other image is captured in `.tif' format. 

The dataset of all the digital images have been collected and the corresponding ground truths have been prepared under the guidance of the expert geologist in our group who is also one of the authors ({\sc tc}) of this paper. More specifically, the ground truth images have been prepared manually under expert supervision by filling the grain portions of a sedimentary rock photomicrograph in white while the rest, i.e., matrix/cement portion is marked in black.

\subsection{Implementation Details} 

\paragraph{\textbf{Pre-processing:}}

We have only seven pairs of images for training purposes. It is challenging to train our {\sc dsgs} network with these limited (i.e., pair of 7) training images shown in Figure~\ref{fig:training}. We apply various pre-processing steps to these training images to generate a larger number of images for training the network. Since the network takes an input image of size $256\times256$, we, first of all, crop $256\times256$ regions from the original image of size $1920 \times 2448$ to generate a large training set. Here, we consider that each image is independent of other for a single pair. 

\begin{itemize}
\item \textbf{Training-Set1:} We crop non-overlapping $256\times256$ regions from the images of the training set containing a pair of each of the seven images and thus create a new training set, named as Training-Set1. The Training-Set1 consists of $1120$ cropped images. Since the original image size, $1920 \times 2448$ is not wholly divisible by the cropped image size $256\times256$, we use padding with $0$ to include the boundary regions into the Training-Set1. 

\item \textbf{Training-Set2:} We crop $256\times256$ regions with $128$ pixels overlapping from the images of the training set to create a new training set, named as Training-Set2. It contains $3950$ cropped images. We use padding with $0$ to include the boundary regions into the Training-Set2.  

\item \textbf{Training-Set3:} Similarly, we crop $256\times256$ regions with $64$ pixels overlapping from the images of the training set to create a new training set, named as Training-Set3. It contains $15149$ cropped images. We use padding with $0$ to include the boundary regions into the Training-Set3. 

\begin{table*}[ht!]
\addtolength{\tabcolsep}{-5.0pt}
\begin{center}
\begin{tabular}{|c|c|c c c c|c c c c|c c c c|c c c c|c c c c|} \hline
\textbf{Method} &\textbf{Training} &\multicolumn{20}{|c|}{\textbf{Quantitative Score on Test Images}} \\ \cline{3-22}
 &\textbf{Dataset} &\multicolumn{4}{|c|}{\textbf{Accuracy}$\uparrow$} &\multicolumn{4}{|c|}{\textbf{Recall}$\uparrow$} &\multicolumn{4}{|c|}{\textbf{Precision}$\uparrow$} &\multicolumn{4}{|c|}{\textbf{F1}$\uparrow$} &\multicolumn{4}{|c|}{\textbf{Jaccard Index}$\uparrow$}  \\ \cline{3-22}
   & &\textbf{Min} &\textbf{Max} &\textbf{Avg} &\textbf{Std} &\textbf{Min} &\textbf{Max} &\textbf{Avg} &\textbf{Std} &\textbf{Min} &\textbf{Max} &\textbf{Avg} &\textbf{Std} &\textbf{Min} &\textbf{Max} &\textbf{Avg} &\textbf{Std} &\textbf{Min} &\textbf{Max} &\textbf{Avg} &\textbf{Std} \\ \hline 
             &Training-Set1  &0.694 &0.882 &0.802 &0.055 &0.686 &0.966 &0.844 &0.093 &0.722 &0.984 &0.857 &0.081 &0.807 &0.905	&0.842 &0.035 &0.405 &0.779 &0.633 &0.113  \\
             &Training-Set2   &0.781 &0.885 &0.829 &0.034 &0.771 &0.944 &0.860 &0.047 &0.815 &0.950 &0.868 &0.041 &0.832 &0.909 &0.862 &0.023 &0.578 &0.781 &0.683 &0.072 \\
             &Training-Set3  &0.708 &0.859 &0.804 &0.045 &0.697 &0.974 &0.881 &0.084 &0.693 &0.985 &0.811 &0.086 &0.799 &0.883 &0.836 &0.027 &0.431 &0.745 &0.643 &0.099  \\ 
{\sc dsgsn}  &Training-Set4  &0.782 &0.861 &0.824 &0.031 &0.791 &0.920 &0.866 &0.051                      &0.764&0.959 &0.851 &0.070 &0.835 &0.867 &0.854 &0.013 &0.581 &0.756 &0.673 &0.074  \\
             &Training-Set5  &0.825 &0.860 &0.842 &0.013 &0.837 &0.930 &0.875 &0.034 &0.784 &0.932 &0.865 &0.057 &0.851 &0.883 &0.868 &0.014 &0.638 &0.754 &0.703 &0.046  \\
             &Training-Set6  &0.826 &0.867 &0.852 &0.016 &0.855 &0.940 &0.886 &0.033 &0.778 &0.920 &0.870 &0.057 &0.852 &0.891 &0.876 &0.015 &0.635 &0.765 &0.718 &0.052 \\ \hline
\end{tabular}
\end{center}
\caption{Shows the performance of the proposed {\sc dsgsn} on Test-Set2 while training with various training sets. Training-Set6 is more effective for training the proposed {\sc dsgsn} for grain segmentation. \textbf{Min:} indicates minimum value, \textbf{Max:} indicates maximum value, \textbf{Avg:} indicates average value, and \textbf{Std:} indicates standard deviation. \label{table_result_various_trainingset}}
\end{table*}

\item \textbf{Test-Set1:} We crop non-overlapping $256\times256$ regions from the images of test set containing four pairs of images, to create a new test set, named as Test-Set1. We apply padding of $0$ to include the boundary regions into the set. The Test-Set1 contains $528$ images.      
\end{itemize}

We create another group of training sets by considering the dependency between images of a single pair in the actual training set containing seven pairs of images. We produce seven new images by averaging the information between each pair of images. Figure~\ref{fig:average} shows the average image in the new training set of each couple of images in the original training set. Similarly, we create a test set containing four images by averaging individual pair of images in the original test set. We then create various training sets by cropping $256\times256$ regions from the images. 

\begin{itemize}
\item \textbf{Training-Set4:} We crop non-overlapping $256\times256$ regions from the images of the training set containing seven average images and create a new training set, named as Training-Set3. The Training-Set3 consists of $560$ cropped images. We use padding with $0$ to include the boundary regions into the Training-Set3. 

\item \textbf{Training-Set5:} We crop $256\times256$ regions with $128$ pixels overlapping from the seven average images to create a new training set, named as Training-Set5. It contains $2030$ cropped images. We use padding with $0$ to include the boundary regions into the Training-Set5.  

\item \textbf{Training-Set6:} Similarly, we crop $256\times256$ regions with $64$ pixels overlapping from the seven average images of the training set to create a new training set, named as Training-Set6. It contains $7784$ cropped images. We use padding with $0$ to include the boundary regions into the Training-Set6. 

\item \textbf{Test-Set2:} We crop non-overlapping $256\times256$ regions from the four average images of test set to create a new test set, named as Test-Set2. We apply padding of $255$ to include the boundary regions into the set. The Test-Set2 contains $288$ images.      
\end{itemize}

\paragraph{\textbf{Training Details:}}

We considered pre-trained ResNet-$18$ model (on ImageNet~\cite{deng2009imagenet}). We do this for all the experiments. We use $3 \times 3$ kernels for all the convolutional layers. The stride of each pooling layer is set to $2$. For the first decomposition block, we use $64$ filters. The number of filters is doubled for each of the subsequent decomposition blocks. The number of filters in each deconvolutional layer is the same as its corresponding pooling layer. We train the model with the training set with batch size $16$ for $60$ epochs. The initial learning rate is set to $0.0005$ with a delay factor of $0.1$. We train our model with the various training sets with the same values of parameters. 

\paragraph{\textbf{Post-processing:}} We merge the segmented outputs of the cropped test images to create the segmented image corresponding to the input image.   

\begin{table*}[ht!]
\begin{center}
\begin{tabular}{|c|c |c|c|c|c|c|} \hline
\textbf{Method} &\textbf{Loss Function} &\multicolumn{5}{|c|}{\textbf{Quantitative Score on Test Images}} \\ \cline{3-7}
  & &\textbf{Accuracy}$\uparrow$ &\textbf{Recall}$\uparrow$ &\textbf{Precision}$\uparrow$ &\textbf{F1}$\uparrow$ &\textbf{Jaccard Index}$\uparrow$  \\ \hline 
{\sc dsgsn} &{\sc bcel} &0.783 &0.819 &0.805 &0.812 &0.638 \\  
{\sc dsgsn} &{\sc wbcel}  &\textbf{0.852} &\textbf{0.886} &\textbf{0.870} &\textbf{0.876} &\textbf{0.718} \\ \hline
\end{tabular}
\end{center}
\caption{Shows the performance of the proposed {\sc dsgsn} with various loss functions on Test-Set2. {\sc bcel:} indicates binary cross entropy loss and {\sc wbcel:} indicates weighted binary cross entropy loss. It highlights that the {\sc wbcel} is more beneficial for grain segmentation to handle the data imbalance problem. \label{table_result_various_loss}}
\end{table*}

\subsection{Evaluation Measures} \label{evaluation_measure}

We use various existing measures such as precision, recall, F1 score, accuracy, Jaccard Index to evaluate the performance of the proposed {\sc dsgsn} on grain segmentation. Precision, recall, and F1 are defined as 

\begin{equation}
\begin{aligned}
precision&=\frac{t_p}{t_p+f_p}, \\ 
recall&=\frac{t_p}{t_p+f_n}, \\
F1&=2*\frac{precision*recall}{precision+recall},
\end{aligned}
\end{equation}
where $t_P$, $f_p$, and $f_n$ are true positive, false positive, and false negative. Accuracy is defined as

\begin{equation}
\begin{aligned}
accuracy&= \frac{c_p}{n_p},
\end{aligned}
\end{equation}
where $c_p$ is the number of correctly predicted pixel and $n_p$ is the total number of pixels. Jaccrad Index is defined as

\begin{equation}
\begin{aligned}
JI&= \frac{\mid A \cap B \mid}{\mid A \cup B \mid},
\end{aligned}
\end{equation}
where $A$ is ground truth segmentation and $B$ is the predicted segmentation output.

\subsection{Ablation Study on Various Training Sets}

The training set consists of $7$ pair of images limited for training {\sc dsgsn}. We use data augmentation approaches to create various training sets consisting of a large number of images from these $7$ pair of images for proper training {\sc dsgsn}. We use overlap cropping (with different overlapping factors) as a data augmentation operation to create various training sets: Training-Set1, Training-Set2, Training-Set3, Training-Set4, Training-Set5, and Training-Set6. Here, we evaluate the performance of {\sc dsgsn} by training with each of the training sets. Table~\ref{table_result_various_trainingset} shows the performance of {\sc dsgsn} trained with different training sets. From the table, we observe that Training-Set6 is effective for training the proposed {\sc dsgsn} on grain segmentation.

\subsection{Ablation Study on Loss Function}

The LinkNet~\cite{linknet} uses cross entropy loss which can not properly handle data imbalance problem. Grain segmentation is a data imbalance problem. Instead of binary cross entropy loss, we proposed weighted binary cross entropy loss defined in Eq.(~\ref{loss_function}) to handle data imbalance problem in grain segmentation. Effectiveness of the proposed weighted binary cross entropy loss is shown in Table~\ref{table_result_various_loss}. We observe from the table that weighted loss better handles the data imbalance problem and improves ($7\%$) segmentation results.

\subsection{Results Comparison with State-of-the-Art Techniques} 

\paragraph{Quantitative Results:}

\textcolor{blue}{Table~\ref{table_result_quantitative} shows the obtained quantitative results by the used architectures. We observe that the proposed method obtains 14\%, 11\%, 8\%, and 0.8\% better than {\sc fcn}, {\sc s}eg{\sc n}et, {\sc u-n}et, and {\sc u-n}et++ (while loss function consider binary cross entropy) with respect to Jaccard Index (see upper part of the Table~\ref{table_result_quantitative}). While we change the loss function binary cross entropy with weighted binary cross entropy in every considered networks, the obtained results are presented in lower part of the Table~\ref{table_result_quantitative}. In this case also, we observe that the proposed method obtains $10\%$, $6\%$, and $3\%$ better results than {\sc fcn}, {\sc s}eg{\sc n}et, and {\sc u-n}et with respect to Jaccard Index. {\sc u-n}et++ obtains the best Jaccard Index ($0.789$) by taking advantage of weighted binary cross entropy loss function with lesser number of parameters ($9.04$M). The proposed {\sc dsgsn} takes ResNet-$18$ as an encoder which is also quite light and is able to produce results comparable with {\sc u-n}et++. From the experimental results, we conclude that the proposed {\sc dsgsn} is better than the existing networks: {\sc fcn}, {\sc s}eg{\sc n}et, {\sc u-n}et, and {\sc u-n}et++. While changing loss function in existing networks, in this case also, the proposed {\sc dsgsn} is better than the existing networks excepting {\sc u-n}et++.}     
\begin{table}[htp!]
\addtolength{\tabcolsep}{-5.5pt}
\begin{center}
\begin{tabular}{|c|c|c|c|c|c|c|} \hline
\textbf{Method} &\textbf{Loss}&\multicolumn{5}{|c|}{\textbf{Quantitative Score on Test Images}} \\ \cline{3-7}
  &\textbf{Function} &\textbf{Accuracy}$\uparrow$ &\textbf{Recall}$\uparrow$ &\textbf{Precision}$\uparrow$ &\textbf{F1}$\uparrow$ &\textbf{JI}$\uparrow$  \\ \hline
{\sc fcn}           &BCEL &0.719  &0.741 &0.683  &0.718  &0.570     \\
{\sc s}eg{\sc n}et  &BCEL &0.740  &0.762  &0.724 &0.743  &0.601  \\
{\sc u-n}et         &BCEL &0.793  &0.809  &0.793   &0.801  &0.638  \\
{\sc u-n}et++       &BCEL &0.832  &0.865  &0.833  &0.846 &0.710  \\ \hline 
{\sc fcn}           &WBCEL &0.762  &0.784 &0.726  &0.769   &0.613     \\
{\sc s}eg{\sc n}et  &WBCEL &0.793  &0.815  &0.777   &0.796  &0.654  \\
{\sc u-n}et         &WBCEL &0.837  &0.853  &0.837   &0.845  &0.682  \\
{\sc u-n}et++       &WBCEL &0.889  &0.919  &0.890    &0.903 &0.789  \\
{\sc dsgsn}         &WBCEL &0.852 &0.886 &0.870 &0.876 &0.718 \\ \hline
\end{tabular}
\end{center}
\caption{Shows the performance comparison of the proposed {\sc dsgsn} with state-of-the-art segmentation networks on Test-Set2.{\sc bcel:} indicates binary cross entropy loss, {\sc wbcel:} indicates weighted binary cross entropy loss, and {\sc ji:} indicates Jacard Index. \label{table_result_quantitative}}
\end{table}

\paragraph{Qualitative Results:} 

\begin{figure*}[htp!]
\centerline{
\tcbox[sharp corners, size = tight, boxrule=0.2mm, colframe=black, colback=white]{
\psfig{figure=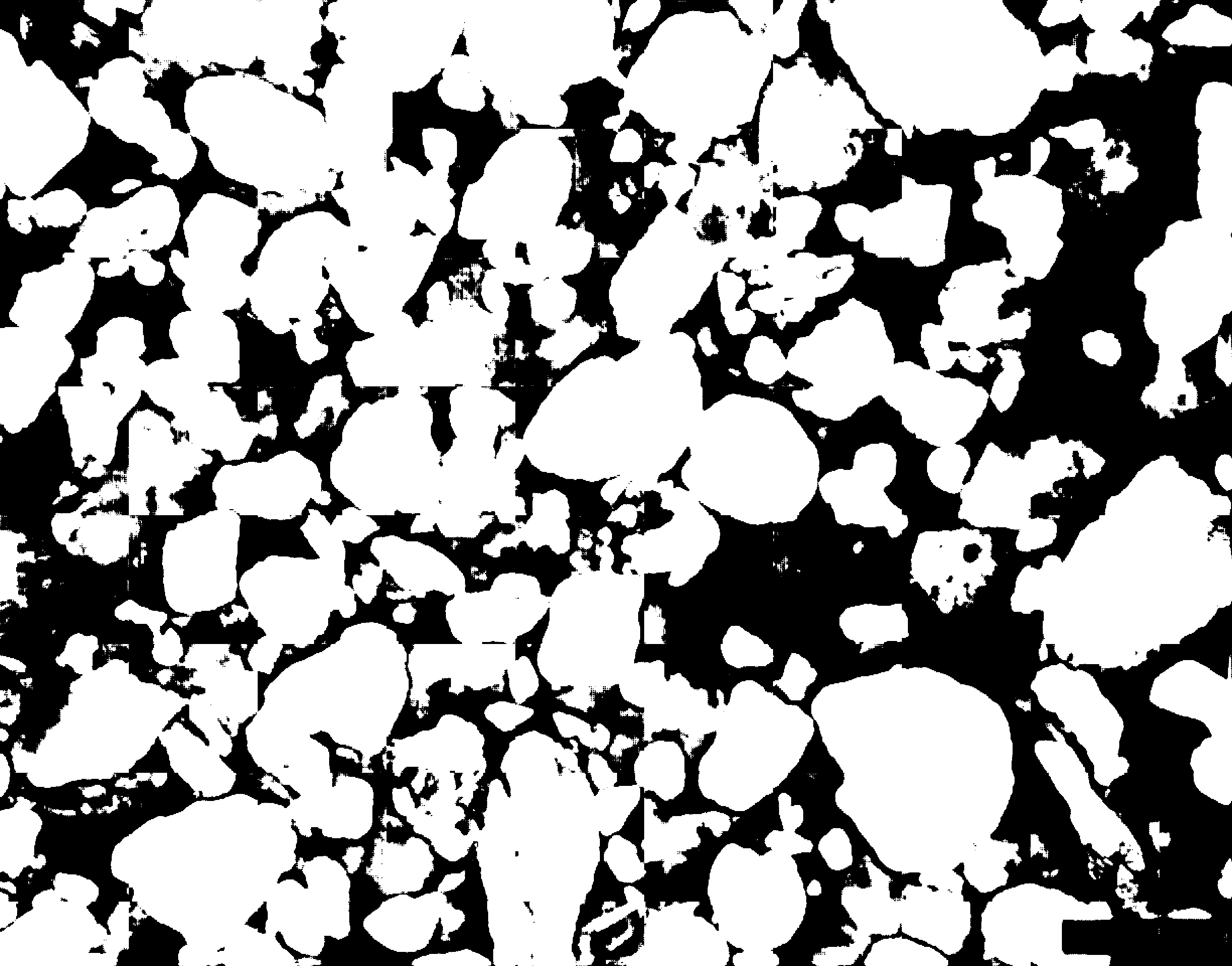,width=0.23\textwidth}}
\hspace{0.001\textwidth}
\tcbox[sharp corners, size = tight, boxrule=0.2mm, colframe=black, colback=white]{
\psfig{figure=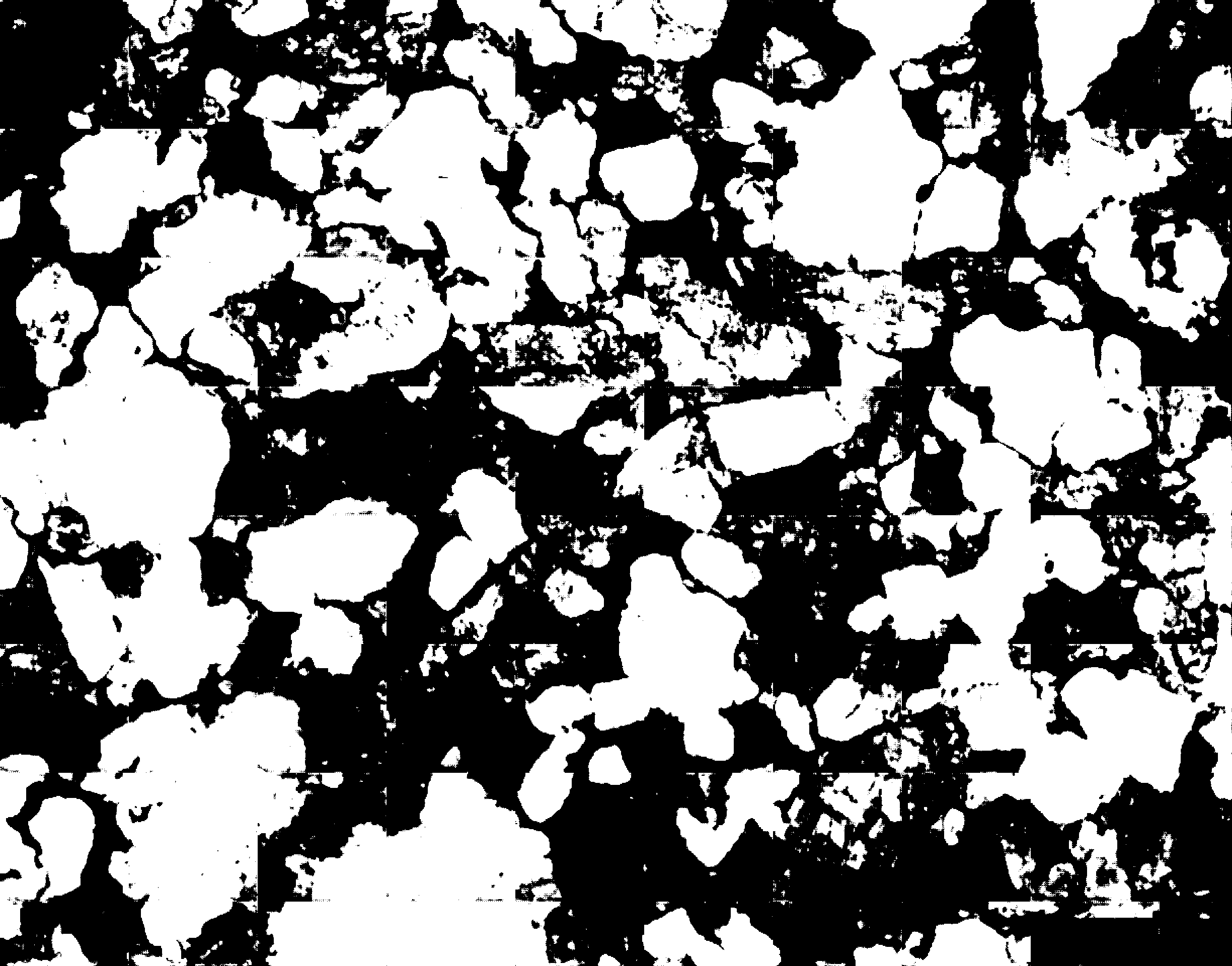,width=0.23\textwidth}}
\hspace{0.001\textwidth}
\tcbox[sharp corners, size = tight, boxrule=0.2mm, colframe=black, colback=white]{
\psfig{figure=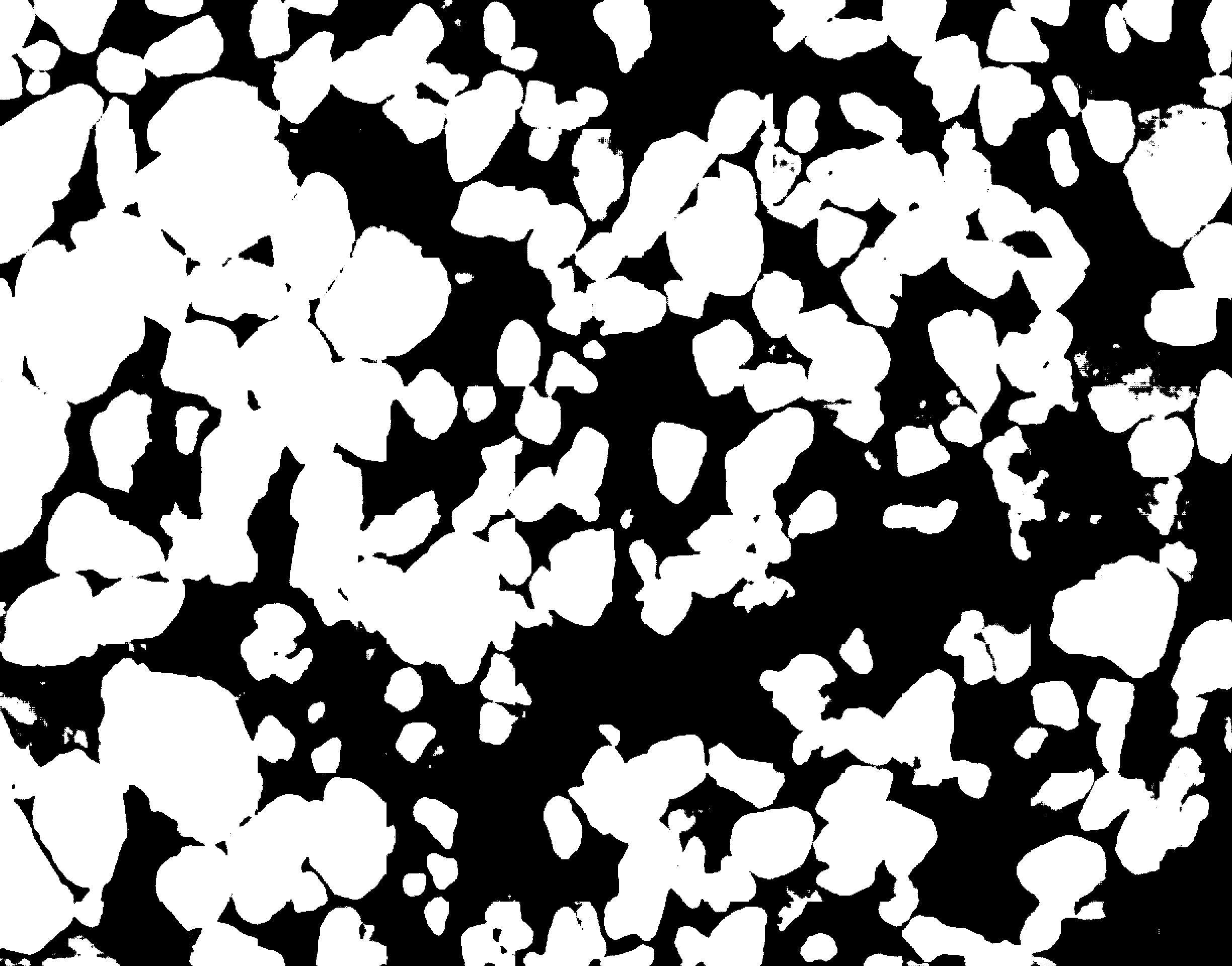,width=0.23\textwidth}}
\hspace{0.001\textwidth}
\tcbox[sharp corners, size = tight, boxrule=0.2mm, colframe=black, colback=white]{
\psfig{figure=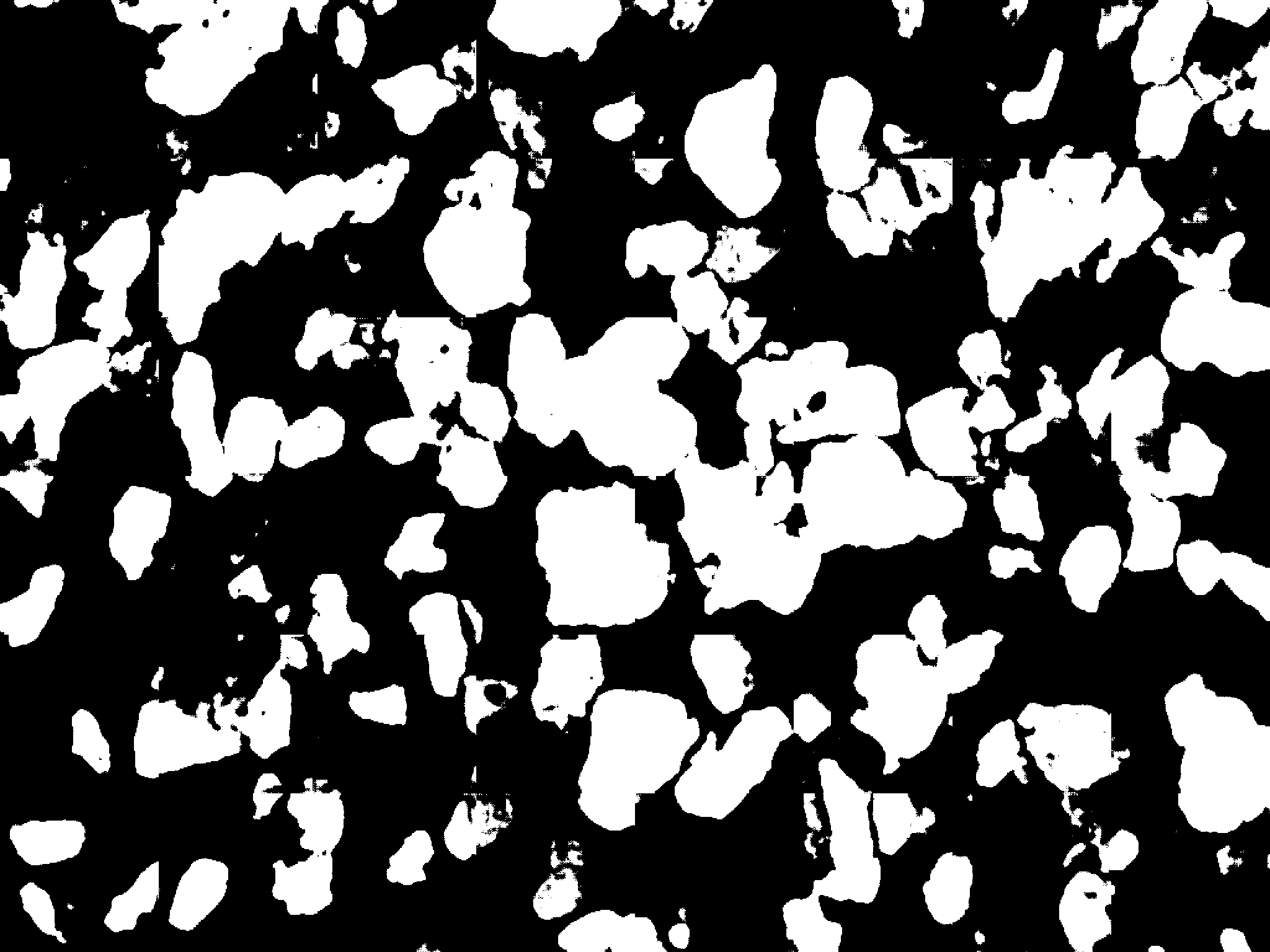,width=0.23\textwidth, height=0.18\textwidth}}}
\centerline{Segmented Results Obtained using {\sc fcn}}
\centerline{
\tcbox[sharp corners, size = tight, boxrule=0.2mm, colframe=black, colback=white]{
\psfig{figure=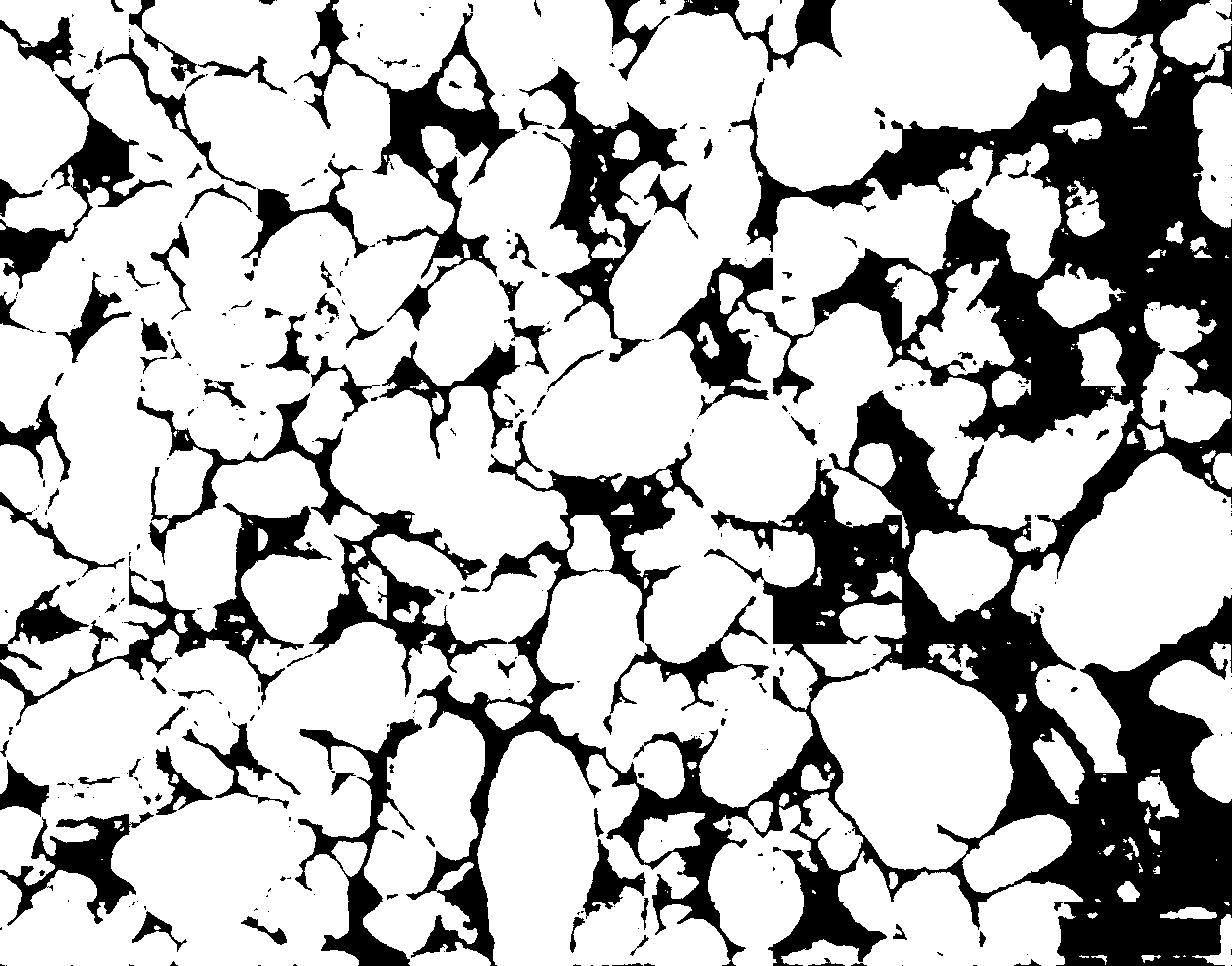,width=0.23\textwidth}}
\hspace{0.001\textwidth}
\tcbox[sharp corners, size = tight, boxrule=0.2mm, colframe=black, colback=white]{
\psfig{figure=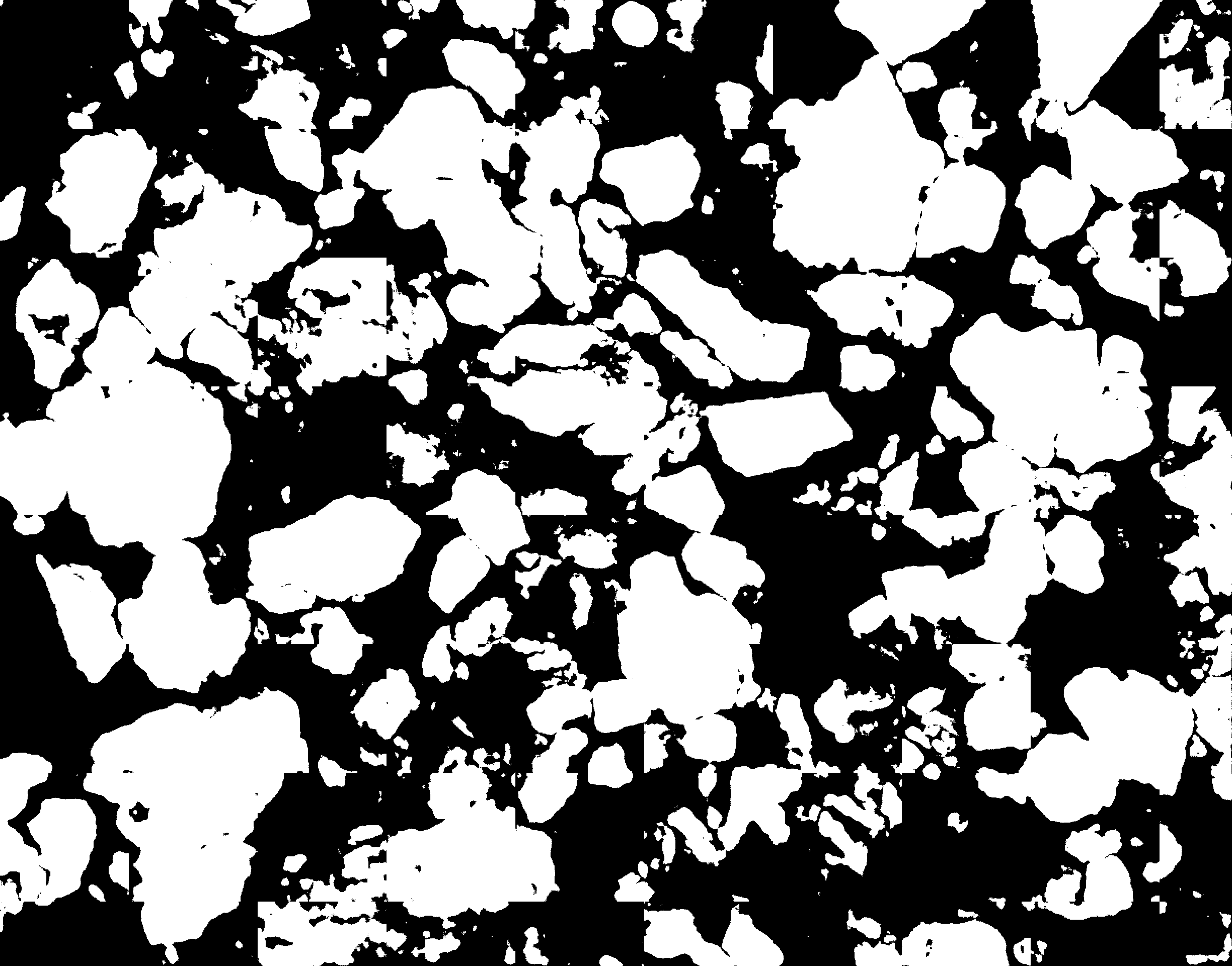,width=0.23\textwidth}}
\hspace{0.001\textwidth}
\tcbox[sharp corners, size = tight, boxrule=0.2mm, colframe=black, colback=white]{
\psfig{figure=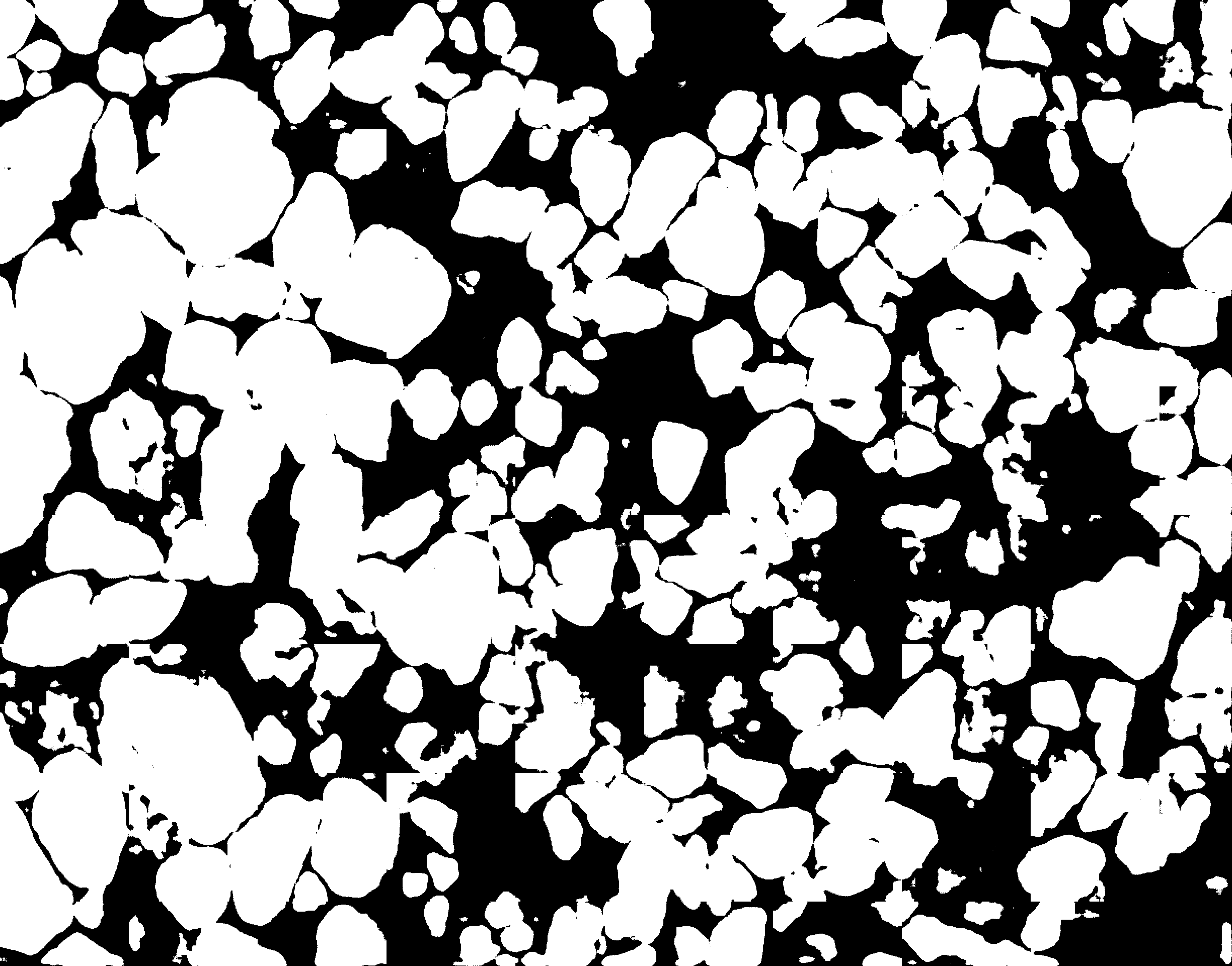,width=0.23\textwidth}}
\hspace{0.001\textwidth}
\tcbox[sharp corners, size = tight, boxrule=0.2mm, colframe=black, colback=white]{
\psfig{figure=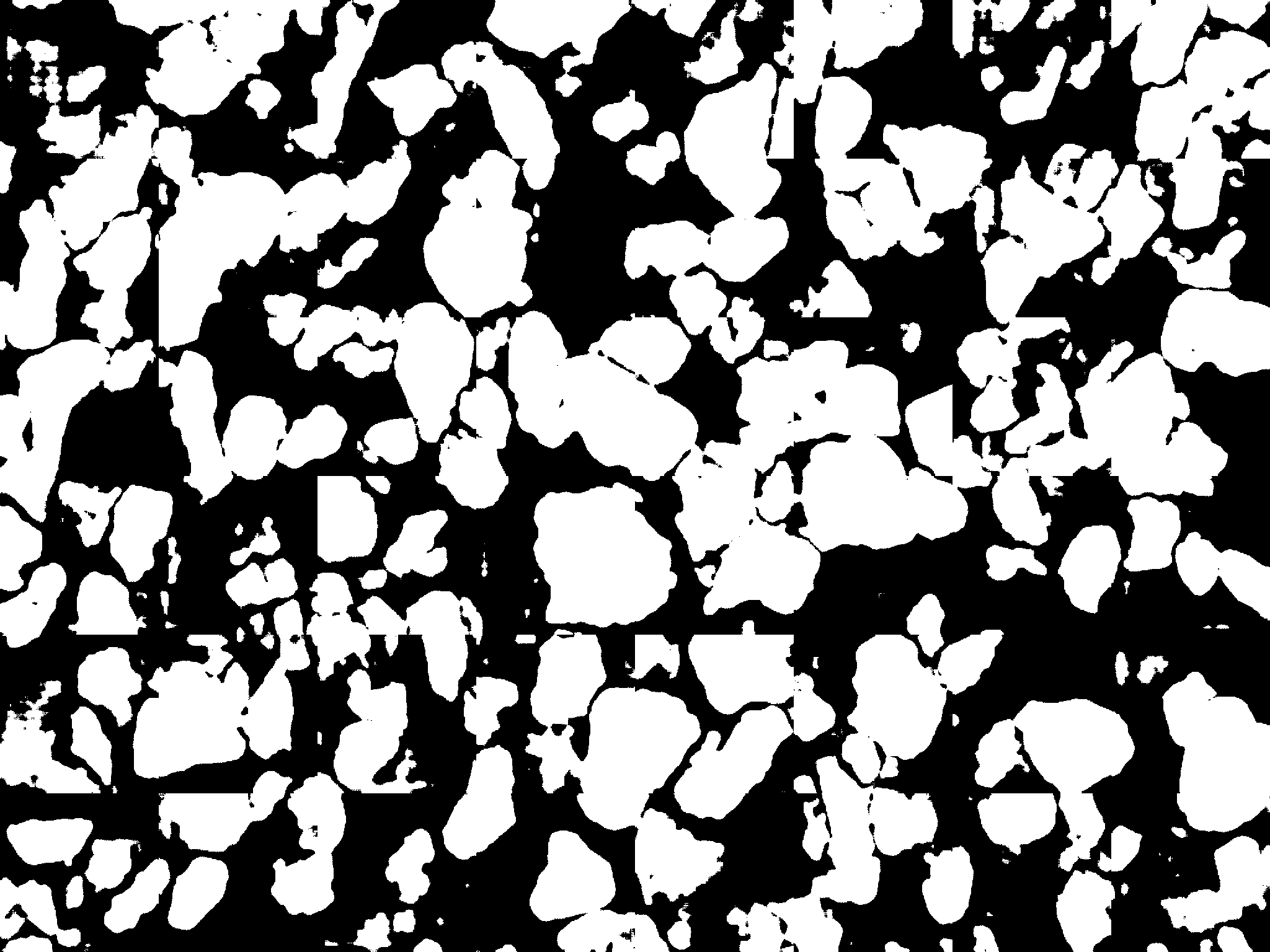,width=0.23\textwidth,height=0.18\textwidth}}}
\centerline{Segmented Results Obtained using {\sc s}eg{\sc n}et}
\centerline{
\tcbox[sharp corners, size = tight, boxrule=0.2mm, colframe=black, colback=white]{
\psfig{figure=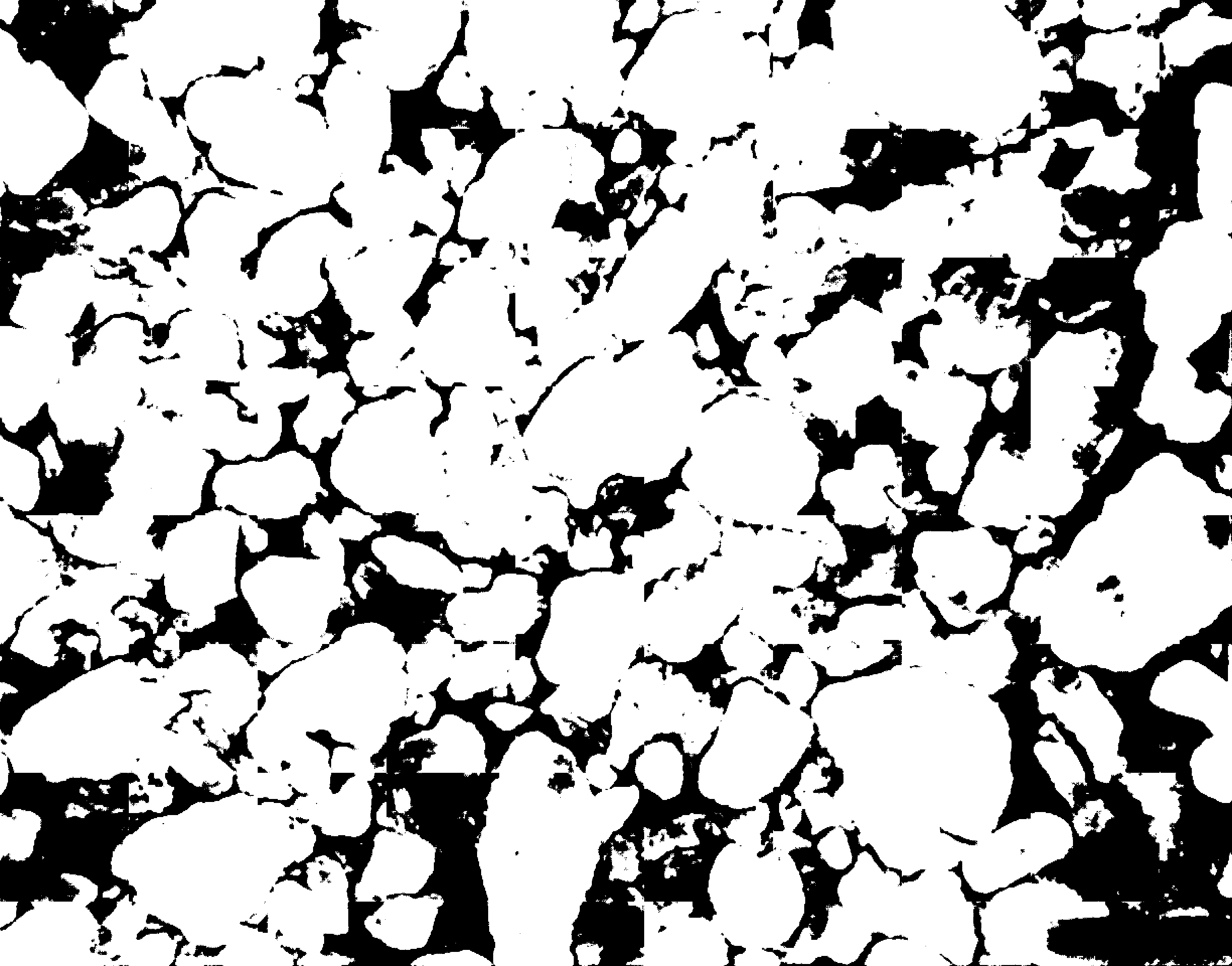,width=0.23\textwidth}}
\hspace{0.001\textwidth}
\tcbox[sharp corners, size = tight, boxrule=0.2mm, colframe=black, colback=white]{
\psfig{figure=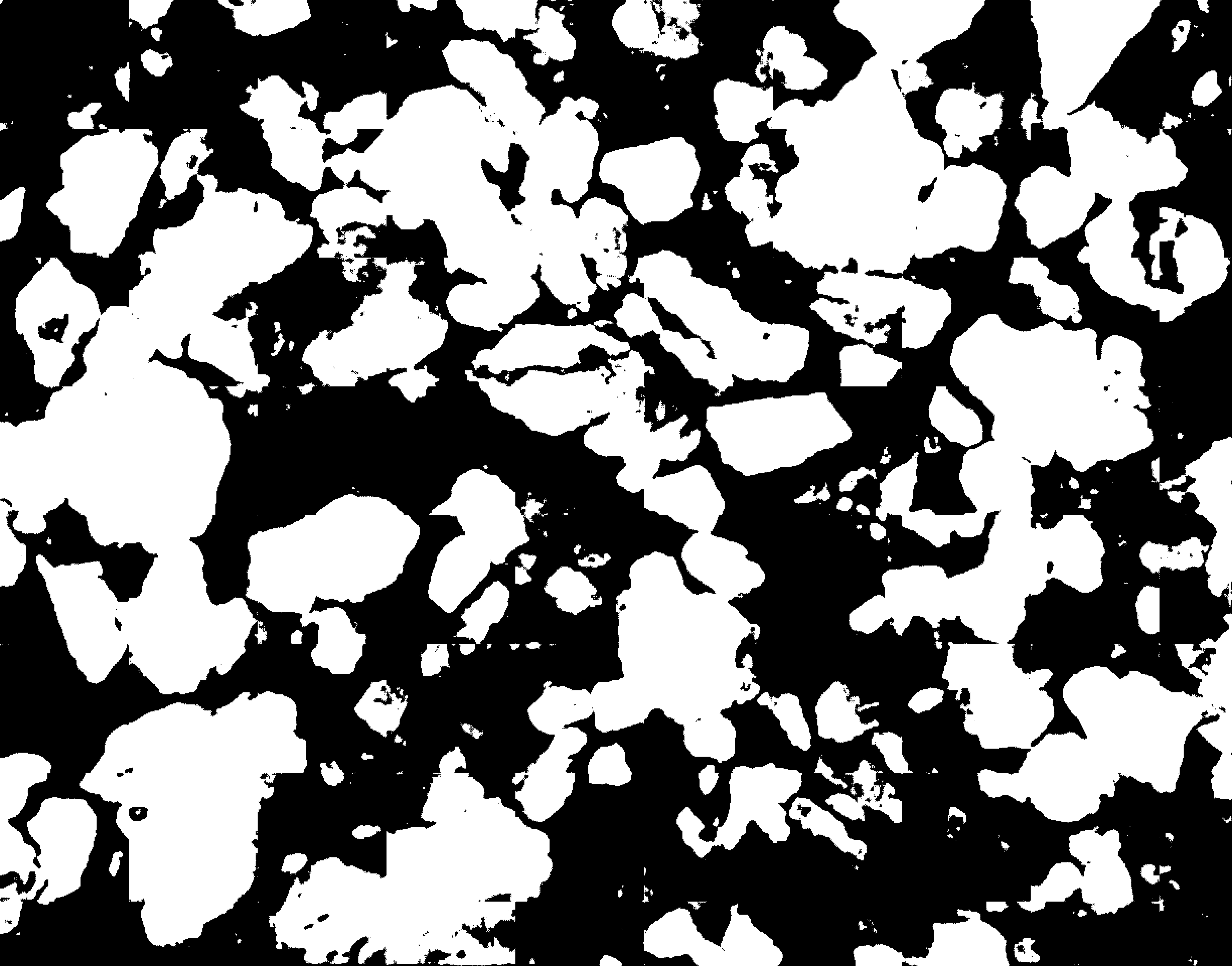,width=0.23\textwidth}}
\hspace{0.001\textwidth}
\tcbox[sharp corners, size = tight, boxrule=0.2mm, colframe=black, colback=white]{
\psfig{figure=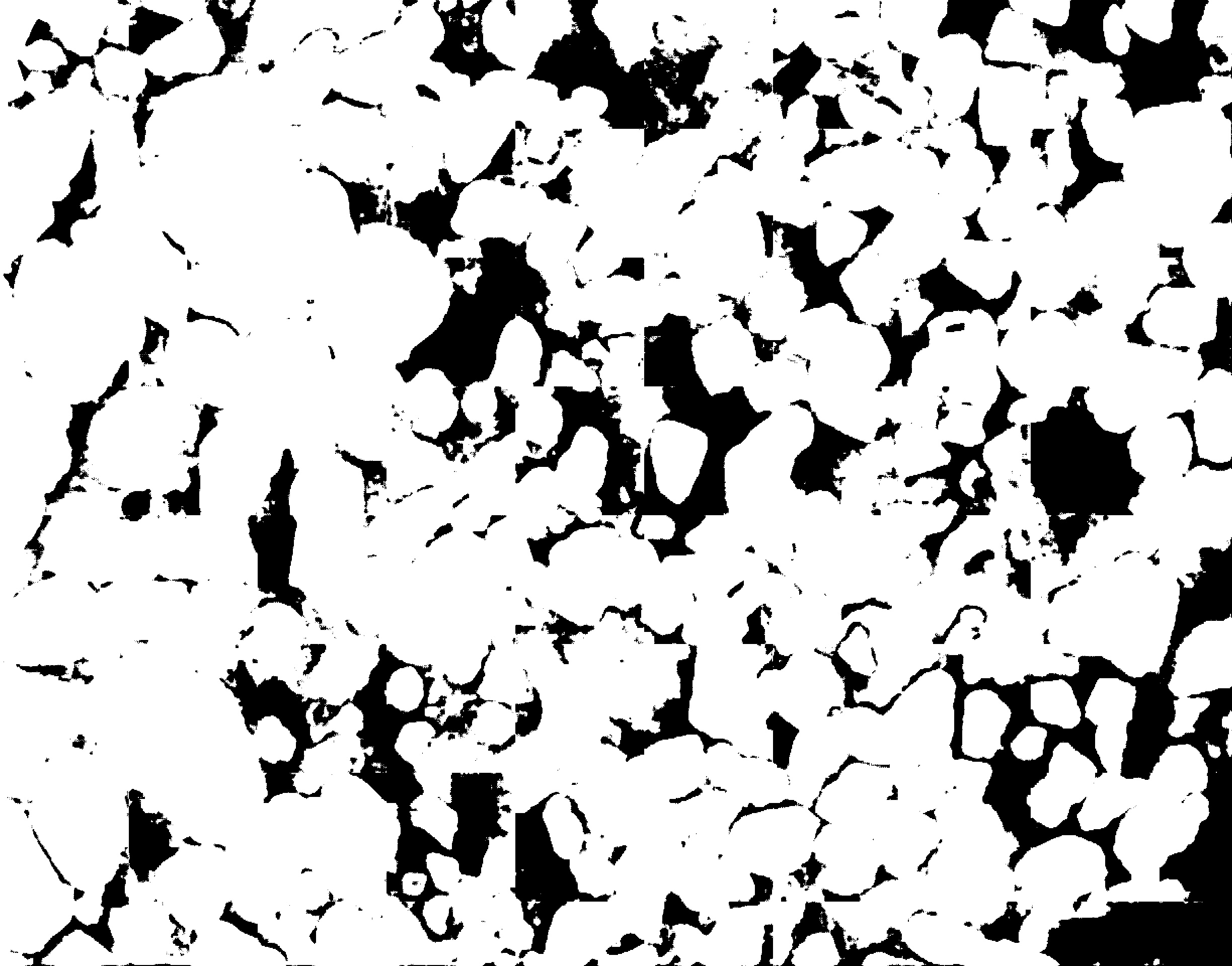,width=0.23\textwidth}}
\hspace{0.001\textwidth}
\tcbox[sharp corners, size = tight, boxrule=0.2mm, colframe=black, colback=white]{
\psfig{figure=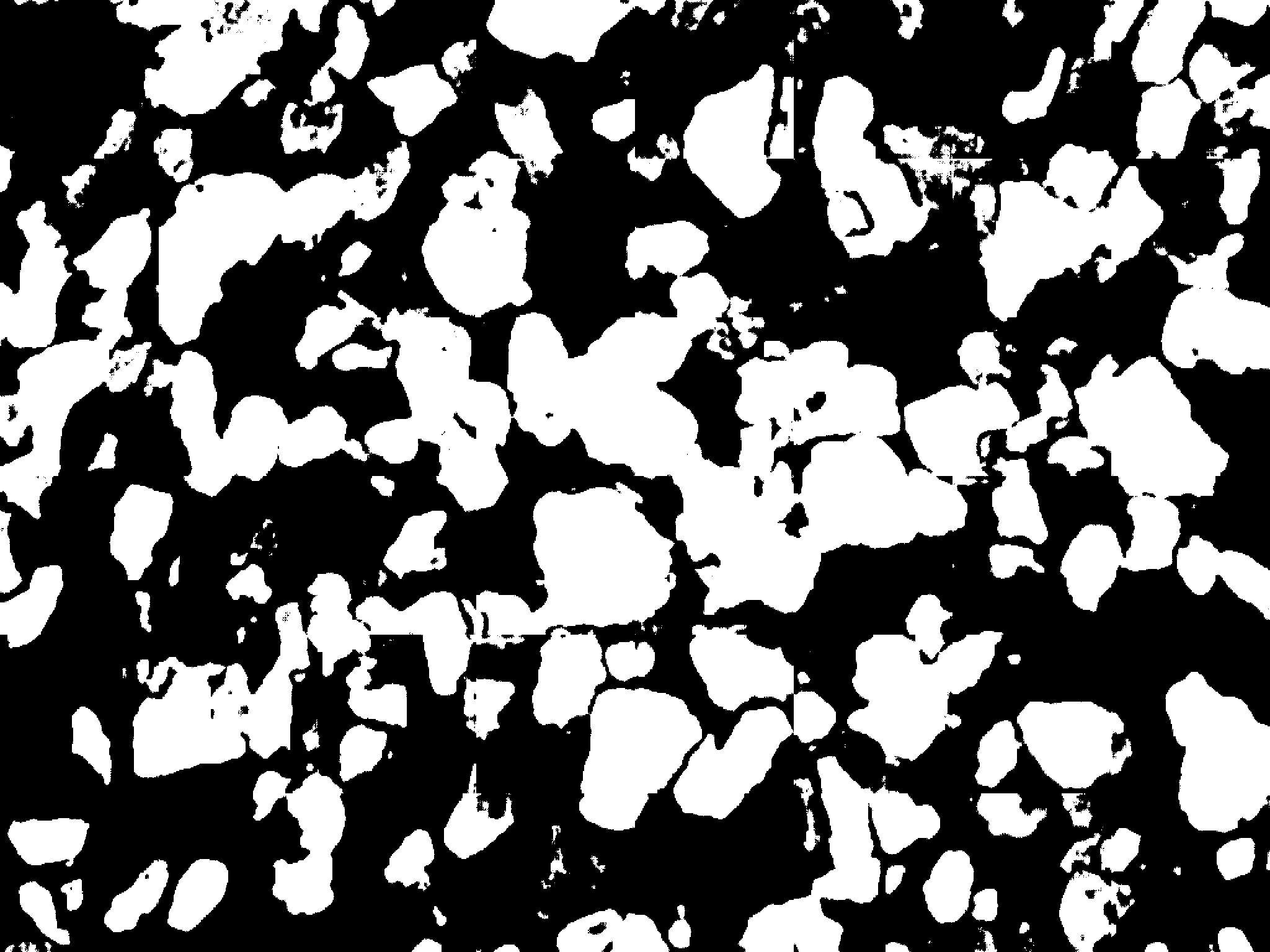,width=0.23\textwidth,height=0.18\textwidth}}}
\centerline{Segmented Results Obtained using {\sc u-n}et}
\centerline{
\tcbox[sharp corners, size = tight, boxrule=0.2mm, colframe=black, colback=white]{
\psfig{figure=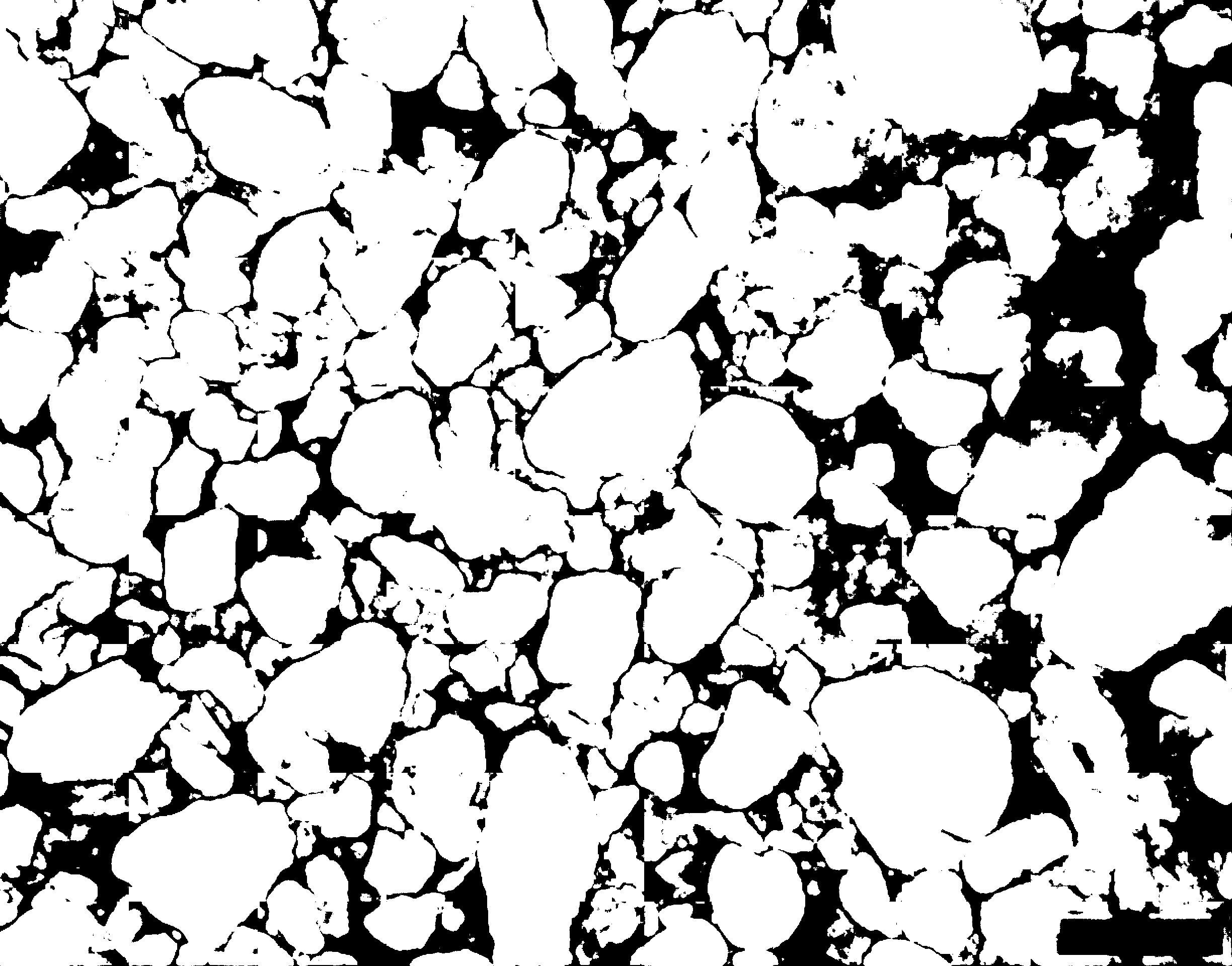,width=0.23\textwidth}}
\hspace{0.001\textwidth}
\tcbox[sharp corners, size = tight, boxrule=0.2mm, colframe=black, colback=white]{
\psfig{figure=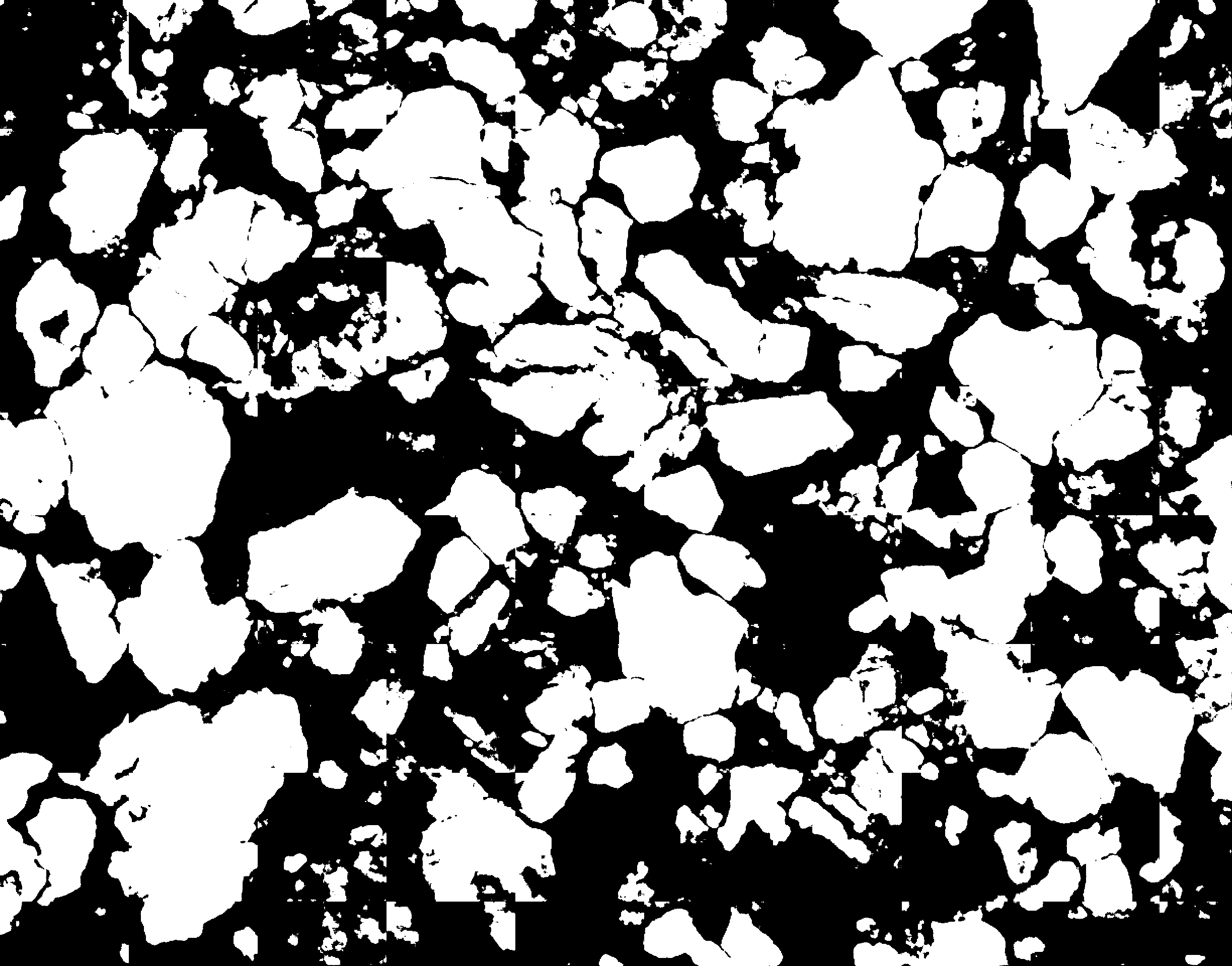,width=0.23\textwidth}}
\hspace{0.001\textwidth}
\tcbox[sharp corners, size = tight, boxrule=0.2mm, colframe=black, colback=white]{
\psfig{figure=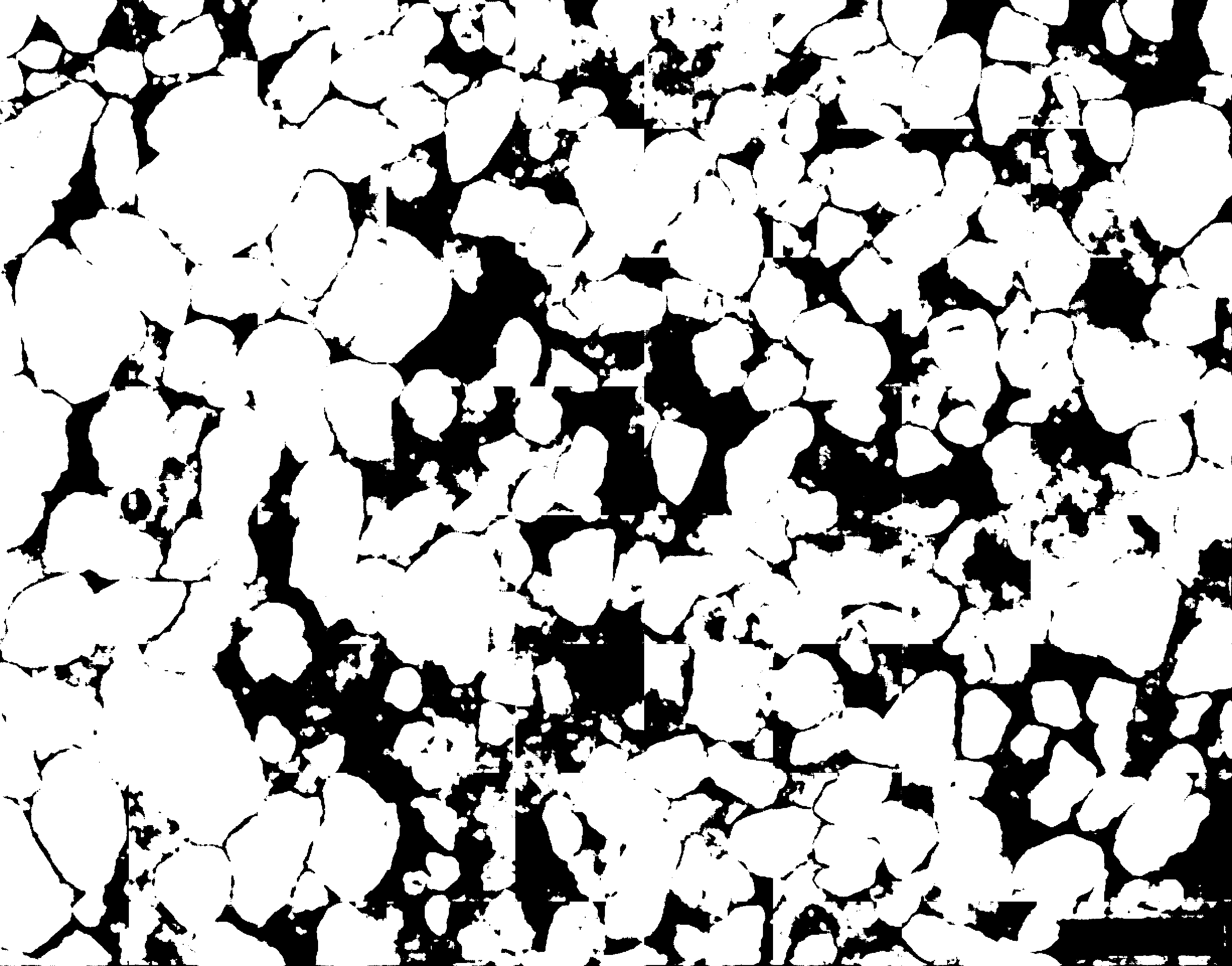,width=0.23\textwidth}}
\hspace{0.001\textwidth}
\tcbox[sharp corners, size = tight, boxrule=0.2mm, colframe=black, colback=white]{
\psfig{figure=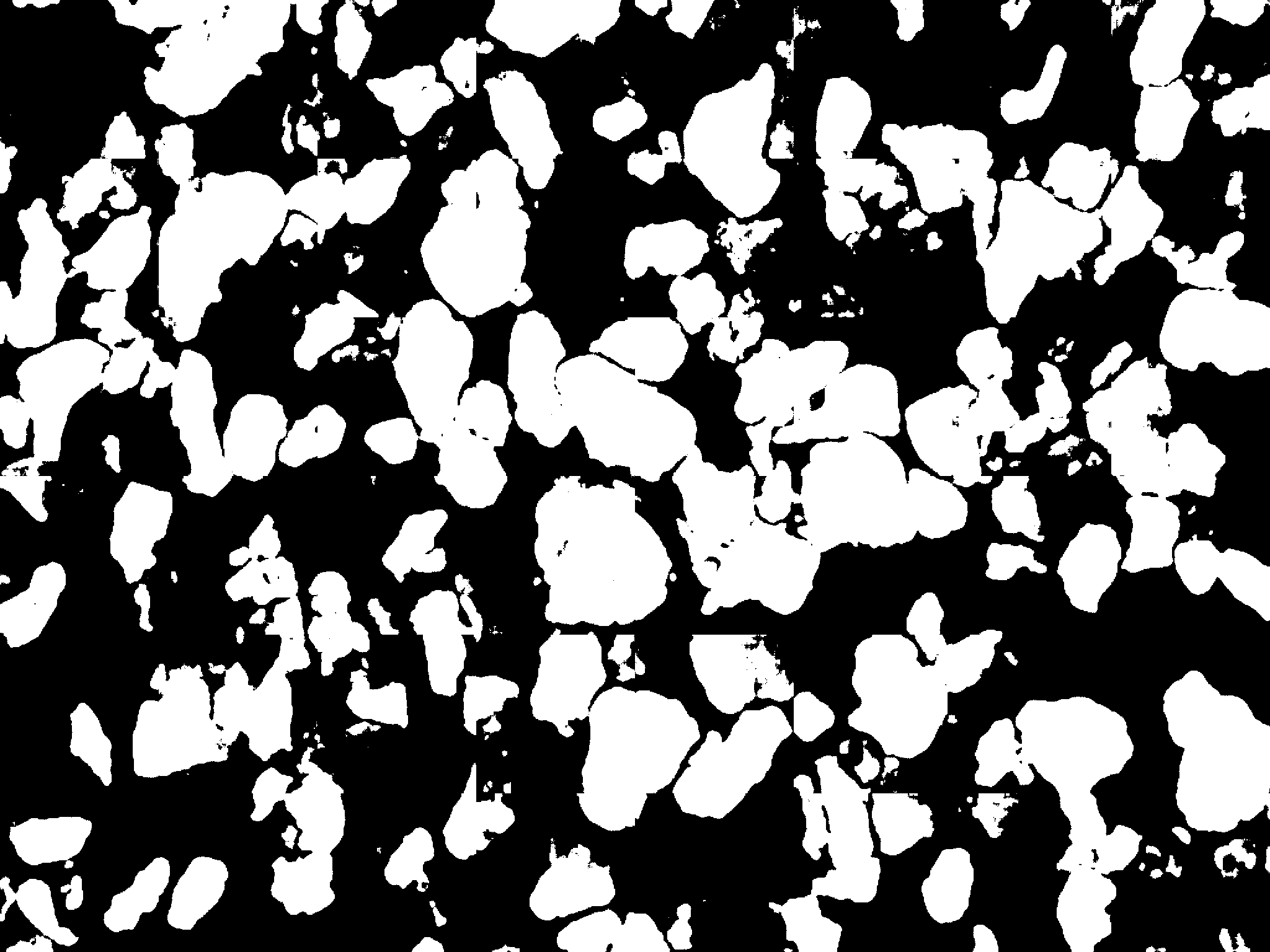,width=0.23\textwidth,height=0.18\textwidth}}}
\centerline{Segmented Results Obtained using {\sc u-n}et++}
\centerline{
\tcbox[sharp corners, size = tight, boxrule=0.2mm, colframe=black, colback=white]{
\psfig{figure=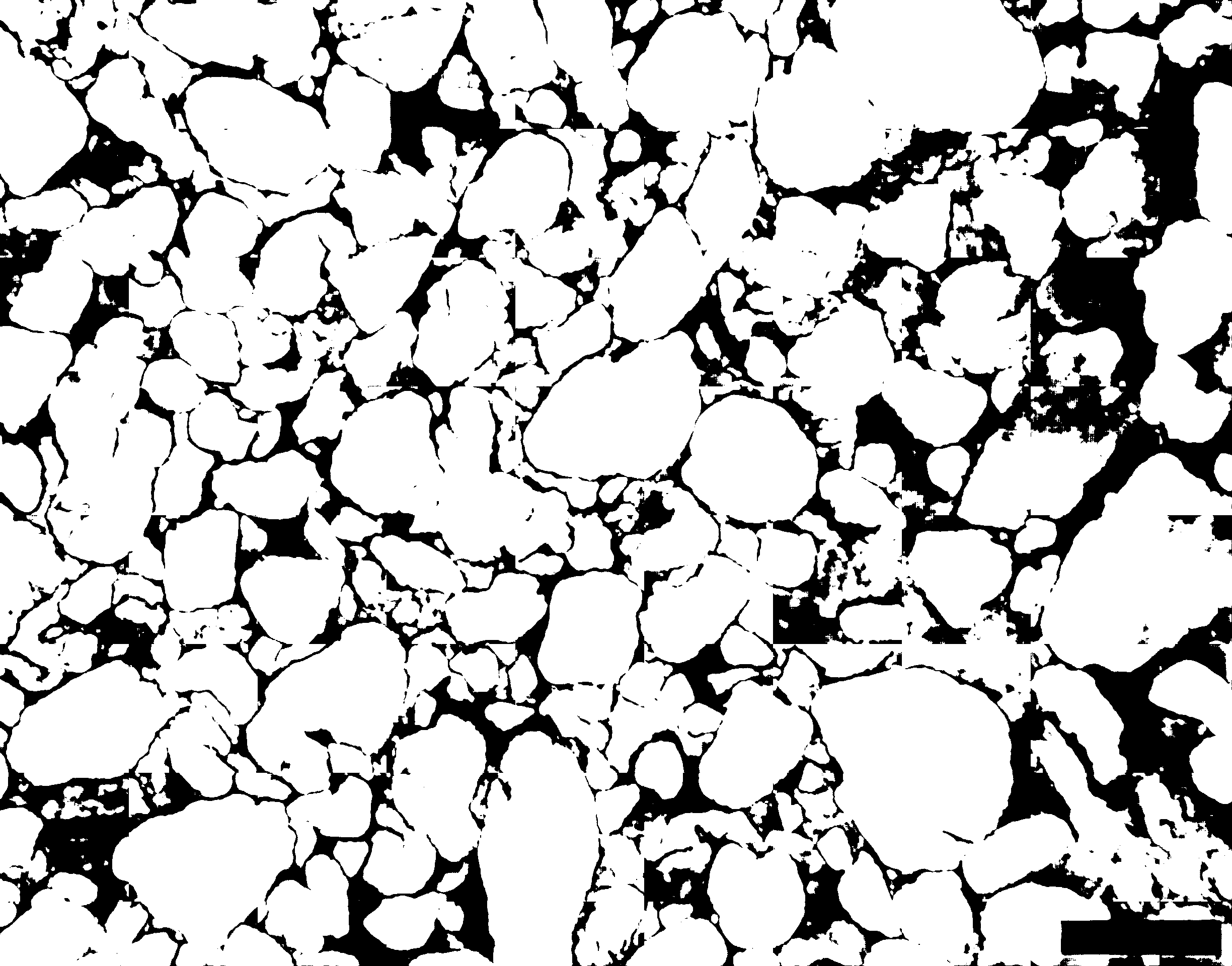,width=0.23\textwidth}}
\hspace{0.001\textwidth}
\tcbox[sharp corners, size = tight, boxrule=0.2mm, colframe=black, colback=white]{
\psfig{figure=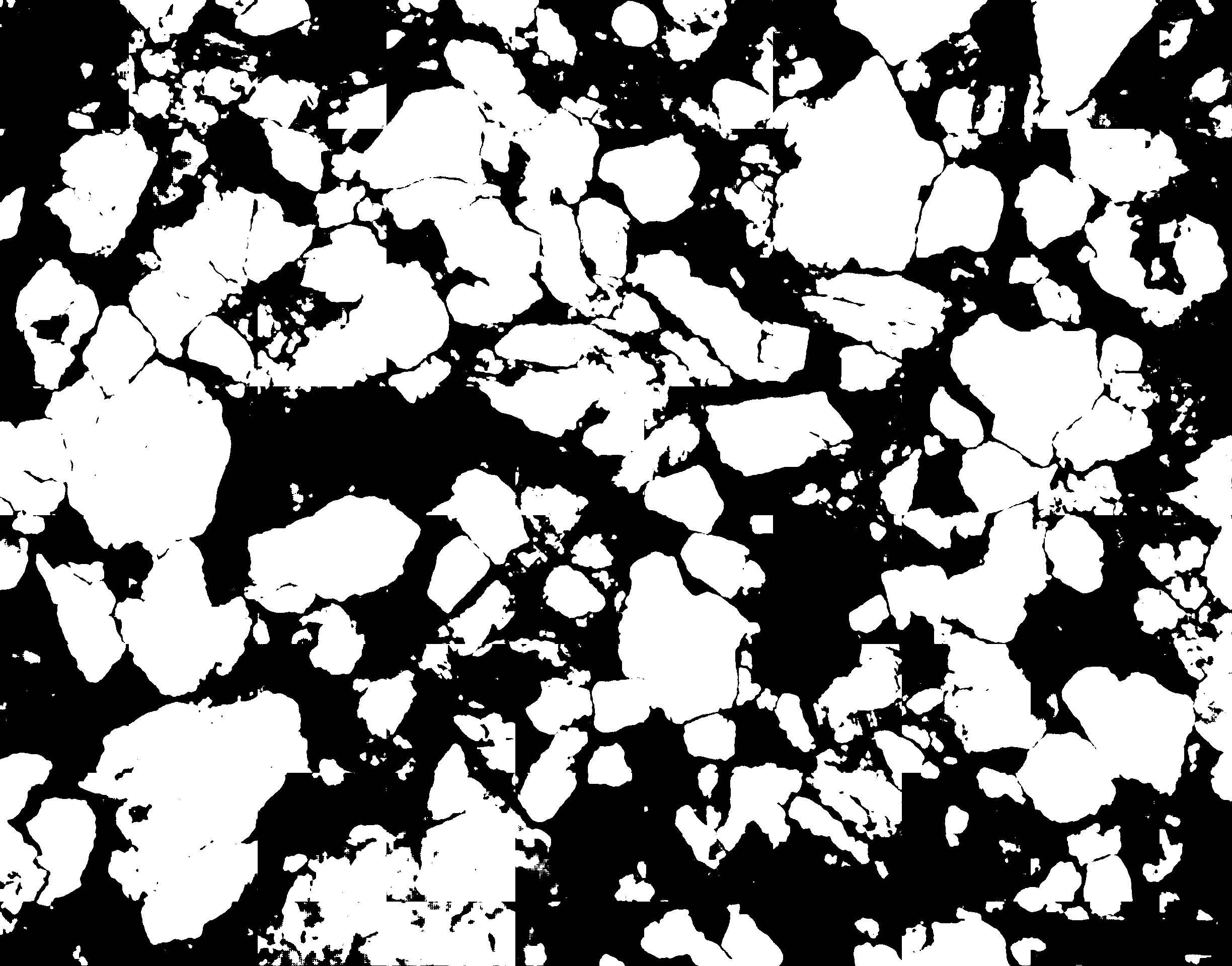,width=0.23\textwidth}}
\hspace{0.001\textwidth}
\tcbox[sharp corners, size = tight, boxrule=0.2mm, colframe=black, colback=white]{
\psfig{figure=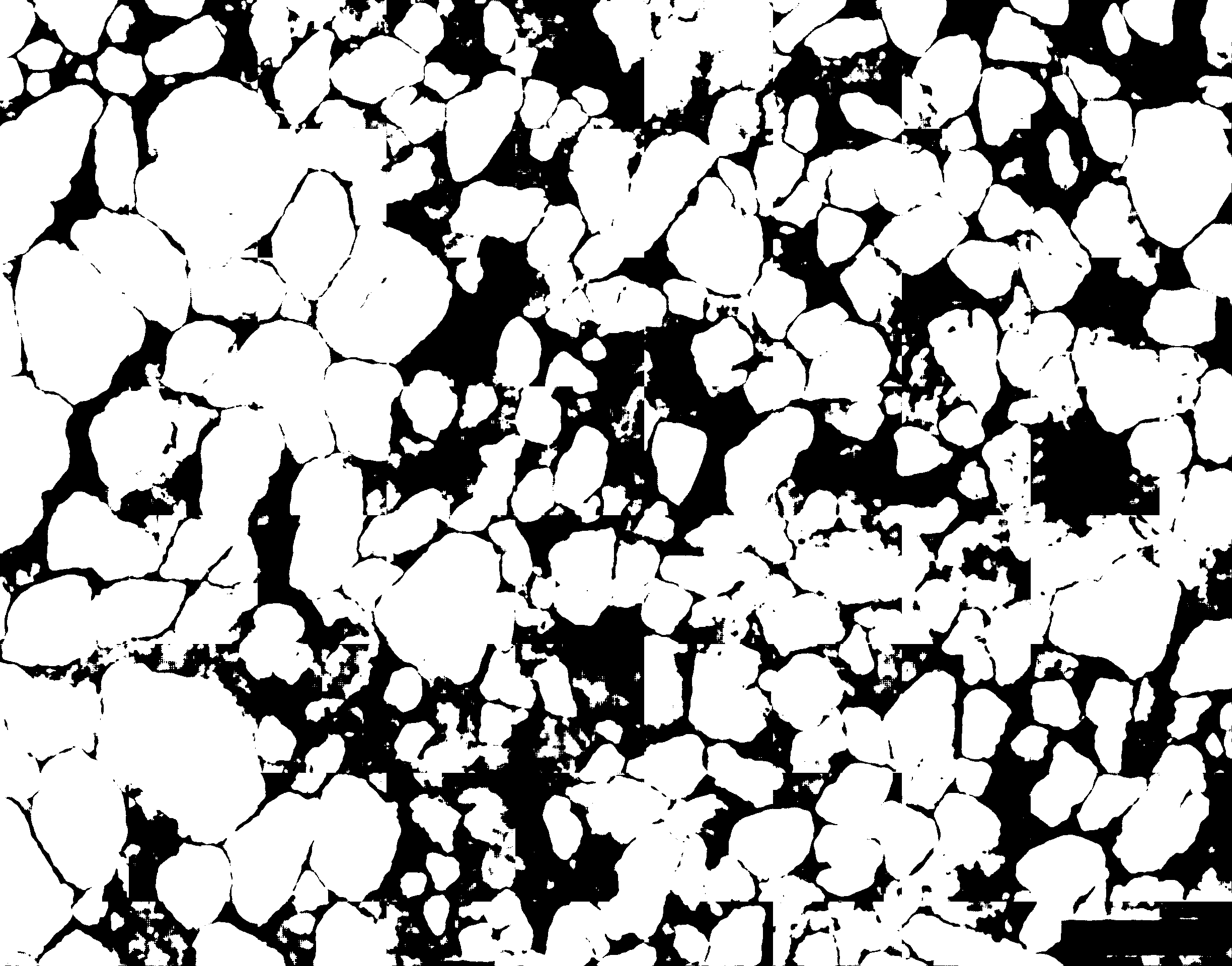,width=0.23\textwidth}}
\hspace{0.001\textwidth}
\tcbox[sharp corners, size = tight, boxrule=0.2mm, colframe=black, colback=white]{
\psfig{figure=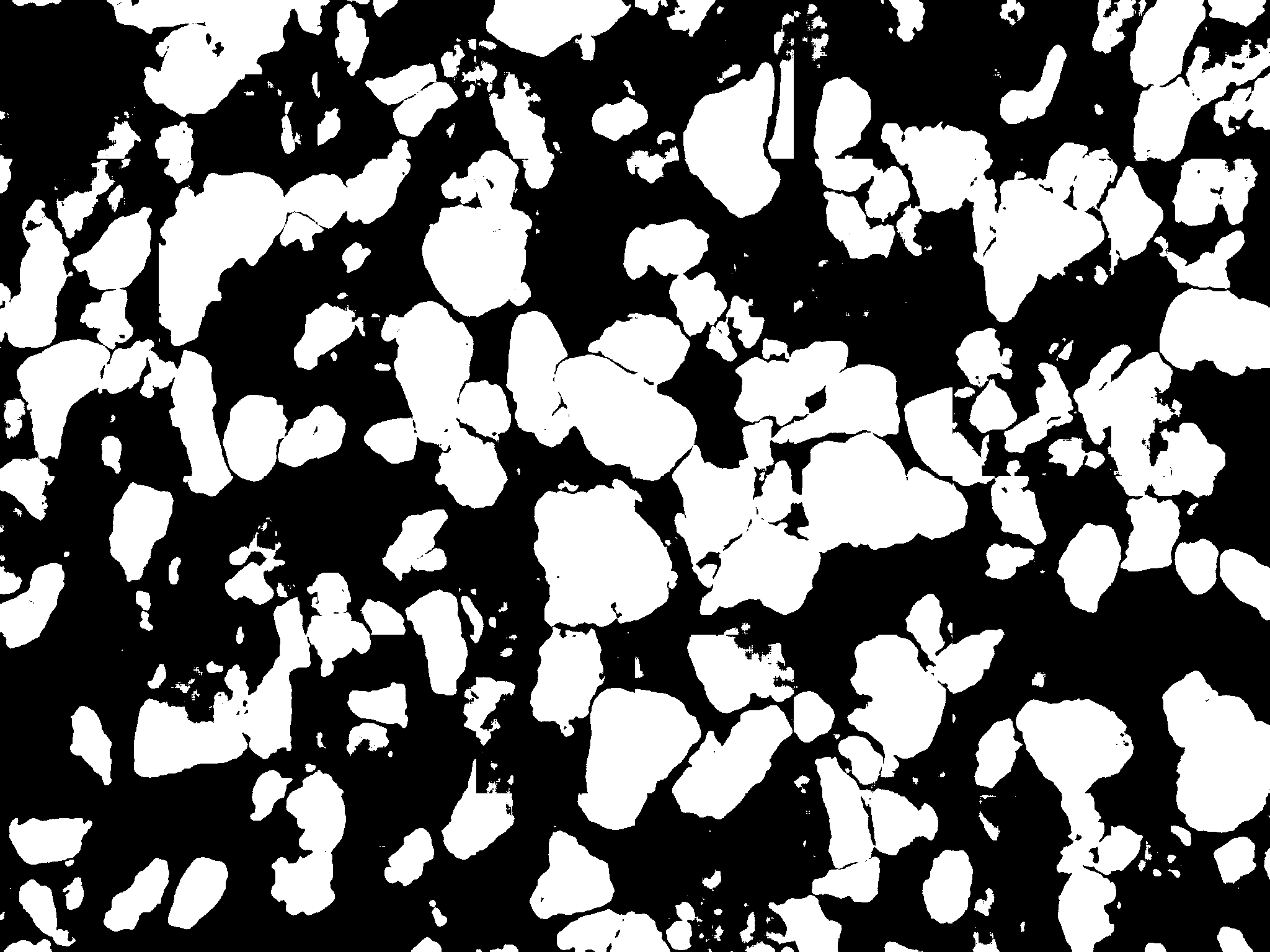,width=0.23\textwidth,height=0.18\textwidth}}}
\centerline{Segmented Results Obtained using {\sc dsgsn}}
\caption{Illustrates qualitative results obtained using various segmentation architectures.  \label{fig:result}}
\end{figure*}

Segmentation results obtained using various considered segmentation architectures are presented in Figure~\ref{fig:result}. From the figure, we observe that {\sc fcn} merges several small foreground patches into a bigger foreground patch. It is also unable to segment tiny foreground patches (containing $50$-$200$ pixels). {\sc s}eg{\sc n}et produces better results than {\sc fcn}, but it also unable to preserve the original structures of the foreground patches. {\sc u-n}et tries to preserve the structures of foreground patches through the encoder-decoder architectures, and input of each encoder layer is also bypassed to the output of its corresponding decoder. It produces better results than {\sc fcn} and {\sc s}eg{\sc n}et. {\sc u-n}et++ obtains the best results among all the networks. While our proposed method {\sc dsgsn} is the second best. Both {\sc u-n}et++ and {\sc dsgsn} preserve structures of foreground patches by properly segmenting pixels in boundary regions.        

\begin{table}[htp!]
\begin{center}
\begin{tabular}{|c|r|} \hline
\textbf{Method} &\textbf{Parameters}  \\ \hline 
{\sc fcn}    &132.00M     \\
{\sc s}eg{\sc n}et  &29.5.0M  \\
{\sc u-n}et         &30.00M \\
{\sc u-n}et++       &9.04M   \\
{\sc dsgsn}  &11.50M \\ \hline
\end{tabular}
\end{center}
\caption{Shows the comparison of the proposed {\sc dsgsn} with the various architectures with respect to parameters. \label{table_parameter}}
\end{table}

\subsection{Parameter Comparison}

Table~\ref{table_parameter} highlights the comparison between the proposed network ({\sc gsdsn}) with existing popular semantic segmentation architectures with respect to their parameters. We observe from the table that {\sc fcn} has the largest set of parameters ($132$M) among our experiments' considered networks. In contrast, {\sc u-n}et++ has the smallest set of parameters ($9.04$M). The proposed network {\sc dsgsn} contains a moderate set of parameters which is considerably lesser than {\sc fcn} and slightly larger than {\sc u-n}et++. Training a network with a large number of parameters also requires a large number of training images. The same is not always be available for real-world problems, e.g., the present case of grain segmentation. It is challenging and cost-intensive for creating training images on grain segmentation. In the case of a lesser amount of training images, our method effectively solves the problem.

\section{Conclusions}\label{conclusion}

This paper presents a deep learning based approach to automatically segment plane and cross-polarized microscopic sandstone images. As distinct from the traditional and existing approaches for this particular problem, we pose the segmentation task as pixel-wise two class classification (semantic segmentation) problem and develop an end-to-end trainable network. The proposed network is data-driven which learns features from the training set during the training and provides a generic solution during the inference time. We also generate a dataset which consists of seven pairs of images with pixel level ground truth annotations for experimental purpose. Due to the  unavailability of sufficient amount of training images for training the proposed network, we use various data augmentation schemes to circumvent the problem. Experiments on this newly generated dataset conclude that the proposed approach leads to better segmented images than various existing segmentation architectures with larger parameters.     

\section*{Compliance with ethical standards}

\subsection*{Conflict of interest}

All authors declare that they have no conflicts of interest.

\subsection*{Ethical approval}

This article does no contain any studies with human participants or animals performed by any of the authors.

\end{document}